\pdfoutput=1

\documentclass[11pt]{article}

\usepackage{acl}

\usepackage{times}
\usepackage{latexsym}
\usepackage{lipsum}  
\usepackage{amsmath, mathtools}  

\usepackage[edges]{forest}
\usetikzlibrary{arrows.meta}
\colorlet{linecol}{black!75}
\usepackage{xkcdcolors} 
\usepackage{tikz}
\usetikzlibrary{backgrounds}
\usetikzlibrary{arrows,shapes}
\usetikzlibrary{tikzmark}
\usetikzlibrary{calc}
\newcommand{\highlight}[2]{\colorbox{#1!17}{$\displaystyle #2$}}

\renewcommand{\highlight}[2]{\colorbox{#1!17}{#2}}

\usepackage[T1]{fontenc}

\usepackage[utf8]{inputenc}

\usepackage{microtype}

\usepackage{inconsolata}

\usepackage{graphicx}
\usepackage{amssymb}

\usepackage{multirow}
\usepackage{booktabs}
\usepackage{arydshln}
\usepackage{float}

\title{ Investigating the translation capabilities of Large Language Models trained on parallel data only}

\author{Javier García Gilabert, Carlos Escolano, Aleix Sant Savall, \\
\textbf{Francesca De Luca Fornaciari, Audrey Mash, Xixian Liao, Maite Melero}\\
Barcelona Super Computing Center (BSC) \\
{\small \tt \{javier.garcia1,carlos.escolano,aleix.santsavall,} \\
{\small \tt francesca.delucafornaciari,audrey.mash,xixian.liao,maite.melero\}}{\small \tt @bsc.es}
}

\definecolor{custom1}{RGB}{15,157,88}
\definecolor{custom2}{RGB}{244, 180, 0}
\definecolor{custom3}{RGB}{66, 133, 244}
\definecolor{custom4}{RGB}{219, 68, 55}
\definecolor{custom5}{RGB}{103,58,183}
\definecolor{custom6}{RGB}{255, 128, 171}

\newcommand{\flores}{\textsc{Flores-200}}
\newcommand{\ntrex}{\textsc{NTREX}}
\newcommand{\bleu}{\textsc{BLEU}}
\newcommand{\comet}{\textsc{COMET}}
\newcommand{\parlam}{\textsc{Plume}}
\newcommand{\nllb}{\textsc{NLLB}}

\usepackage{amssymb}
\usepackage{arydshln}
\makeatletter
\def\adl@drawiv#1#2#3{%
        \hskip.5\tabcolsep
        \xleaders#3{#2.5\@tempdimb #1{1}#2.5\@tempdimb}%
                #2\z@ plus1fil minus1fil\relax
        \hskip.5\tabcolsep}
\newcommand{\cdashlinelr}[1]{%
  \noalign{\vskip 1.3pt
           \global\let\@dashdrawstore\adl@draw
           \global\let\adl@draw\adl@drawiv}
  \cdashline{#1}[.4pt/2pt]
  \noalign{\global\let\adl@draw\@dashdrawstore
           \vskip 3pt}}
\makeatother

\begin{document}
\maketitle
\begin{abstract}

In recent years, Large Language Models (LLMs) have demonstrated exceptional proficiency across a broad spectrum of Natural Language Processing (NLP) tasks, including Machine Translation. However, previous methods predominantly relied on iterative processes such as instruction fine-tuning or continual pre-training, leaving unexplored the challenges of training LLMs solely on parallel data. In this work, we introduce \parlam\ (\textbf{P}arallel \textbf{L}ang\textbf{u}age \textbf{M}od\textbf{e}l), a collection of three 2B LLMs featuring varying vocabulary sizes (32k, 128k, and 256k) trained exclusively on  Catalan-centric parallel examples. These models perform comparably to previous encoder-decoder architectures on 16 supervised translation directions and 56 zero-shot ones. Utilizing this set of models, we conduct a thorough investigation into the translation capabilities of LLMs, probing their performance, the impact of the different elements of the prompt, and their cross-lingual representation space. We release \href{https://huggingface.co/projecte-aina/Plume32k}{ \parlam\ 32k}, \href{https://huggingface.co/projecte-aina/Plume128k}{ \parlam\ 128k} and \href{https://huggingface.co/projecte-aina/Plume256k}{ \parlam\ 256k} checkpoints as well as our \href{https://github.com/projecte-aina/Plume}{training code}. 

\end{abstract}

\section{Introduction}

Neural Machine Translation (NMT) has traditionally relied on encoder-decoder architectures, where an encoder processes the source sentence and a decoder generates the target sentence based on the encoder's output. However, recent advancements have moved away from this paradigm, with the introduction of decoder-only Large Language Models (LLMs). In these models, the source sentence acts as a prompt, eliminating the need for a conventional encoder.

With the rise of LLMs, research has increasingly focused on adapting these models for translation tasks by using techniques such as prompt-tuning \cite{DBLP:conf/icml/0006HB23}, instruction-finetuning \cite{xu2023paradigm}, or continual pretraining \cite{rei-etal-2022-comet}. While these methods have shown impressive results, they open new questions about the performance of LLMs when trained exclusively on parallel data, and therefore, the possibility of having models that are trained directly on the task of machine translation. Additionally, the majority of these models are trained predominantly on English-centric-corpora.

To address these questions, our paper proposes a new approach consisting of training LLMs solely on parallel corpora to evaluate their efficacy in machine translation (MT). Our investigation revolves around questions such as: How does an LLM trained exclusively on parallel data perform? And how does the model leverage prompt information to ensure accurate translations?

Our contributions are twofold: Firstly, we introduce \parlam\ (\textbf{P}arallel \textbf{L}ang\textbf{u}age \textbf{M}od\textbf{e}l), an innovative ensemble comprising three multilingual 2B LLMs, trained from scratch on Catalan-centric parallel data. Each model has a different vocabulary size (32k, 128k and 256k). All models are proficient in 16 supervised translation  directions, as well as 56 zero-shot translation directions. Results show comparable results to previous encoder-decoder architectures of similar size. 

Secondly, to understand how these models work, we study how they utilize contextual information across different layers to execute translation tasks effectively. Our experiments show distinctive attention patterns associated with the different parts of the prompt, and how they vary through the different attention blocks. We also observe how languages use the source tag information differently, leading to a large performance variability when this token is missing. As a byproduct, we propose a strategy to remove attention heads with minimal performance loss. Finally, we study the cross-lingual space learned by the models and how it progresses through the model's attention blocks.


\section{Related work}

Neural Machine Translation (NMT) has predominantly relied on encoder-decoder architectures \cite{DBLP:conf/ssst/ChoMBB14, DBLP:journals/corr/BahdanauCB14, DBLP:conf/nips/SutskeverVL14}. These methods have proven effective by conditioning language models to generate translations that accurately retain the meaning of the source sentence. Moreover, these systems are easily extendable to multilingual scenarios, enabling zero-shot translation between language pairs that have not been seen together during training \cite{DBLP:conf/naacl/FiratCB16, wu2016google}.

Over the years, some approaches to NMT have dropped the traditional encoder-decoder setup to adopt decoder-only architectures \cite{fonollosa2019joint, NEURIPS2018_4fb8a7a2}. Although these methods showed promise, they did not become the standard due to issues with context loss and hallucinations \cite{fu2023decoder}.

Recent advancements in training Large Language Models (LLMs) \cite{touvron2023llama, jiang2023mistral, gemmateam2024gemma, abdin2024phi3}, including techniques like scaling and Rotary Embeddings \cite{DBLP:journals/ijon/SuALPBL24}, have significantly enhanced the ability of decoder-only architectures to handle long contexts of hundreds or even thousands of tokens. Consequently, several studies have proposed leveraging pretrained LLMs for NMT through continual pretraining and instruction tuning \cite{alves2024tower, xu2023paradigm,yang-etal-2023-BigTranslate}. These methods have demonstrated results comparable to traditional encoder-decoder systems, while also supporting multiple translation directions.

However, training and adapting these systems to various languages remains challenging \cite{ali2023tokenizer}. Creating a vocabulary that accurately represents all supported languages can lead to performance disparities of up to 68\% on some downstream tasks. Additionally, interpretability methods have gained popularity in order to understand better how models utilize provided information and to guide further improvements \cite{DBLP:conf/acl/VoitaTMST19, DBLP:conf/emnlp/VoitaST19a, ferrando2024primer}.

\section{Methodology}

\subsection{Catalan-Centric Dataset}  

In order to study zero-shot translation using a decoder-only architecture, we employ a Catalan-centric dataset. This dataset pairs Catalan sentences with their counterparts in one of eight other languages: Spanish, French, Italian, Portuguese, Galician, German, English, and Basque. Specifically, for each language, we include translation directions both to Catalan  (xx$\rightarrow$ca) and from Catalan (ca$\rightarrow$xx). The dataset consists of 783.6M sentences and 30.9 billion words.


\paragraph{Data preprocessing} All data is first filtered using LaBSE \cite{feng-etal-2022-language}. Off-target translations are removed using the \textit{Lingua}\footnote{\url{https://github.com/pemistahl/lingua-py}} library. Following the filtering process, the data undergoes deduplication and punctuation normalization using the \textit{Bifixer} library \cite{prompsit:2020:EAMT}. Further details about the dataset are provided in Appendix \ref{sec:dataset}.




\begin{figure*}[t]
\begin{equation*}
    \tikzmarknode{s}{\highlight{custom1}{\text{<s>}}} \ 
    \tikzmarknode{cat}{\highlight{custom2}{\text{[cat\_Latn]}}} \ 
    \tikzmarknode{source}{\highlight{custom3}{\text{I com és ell? D'on és?}}} \ 
    \tikzmarknode{nspa}{ \highlight{custom4}{\text{ \textbackslash n[spa\_Latn] }} } \ 
    \tikzmarknode{target}{ \highlight{custom5}{ \text{¿Y cómo es él? ¿De dónde es?}  } } \ 
    \tikzmarknode{es}{\highlight{custom6}{\text{</s>}}}
\end{equation*}
\begin{tikzpicture}[overlay, remember picture,>=stealth,nodes={align=center, font=\small, inner ysep=2pt},<-]
    \path (s.north) ++ (0,2em) node[anchor=south west,color=custom1] (sparam){Begin of Sentence (BOS)};
    \draw [color=custom1](s.north) |- ([xshift=-0.3ex,color=red]sparam.south east);
    
    \path (cat.north) ++ (-2.1,-2em) node[anchor=north west,color=custom2] (catparam){Source tag};
    \draw [color=custom2](cat.south) |- ([xshift=-0.3ex,color=red]catparam.south west);
    
    \path (source.south) ++ (0,-0.8em) node[anchor=north west, color=custom3] (sourcelabel){Source sentence};
    \draw [color=custom3](source.south) |- ([xshift=-0.3ex,color=custom3]sourcelabel.south east);
    
    \path (nspa.north) ++ (0,1.8em) node[anchor=south west,color=custom4] (nspaparam){Target tag};
    \draw [color=custom4](nspa.north) |- ([xshift=-0.3ex,color=red]nspaparam.south east);

    \path (target.north) ++ (-2.3,-2em) node[anchor=north west,color=custom5] (targetparam){Target sentence};
    \draw [color=custom5](target.south) |- ([xshift=-0.3ex,color=red]targetparam.south west);

    \path (es.north) ++ (0,1.8em) node[anchor=south east,color=custom6] (esparam){End of sentence (EOS)};
    \draw [color=custom6](es.north) |- ([xshift=-0.3ex,color=red]esparam.south west);
\end{tikzpicture}

\hspace{5 em}
\caption{Prompt strategy used to train \parlam.}\label{sentence_example}
\end{figure*}
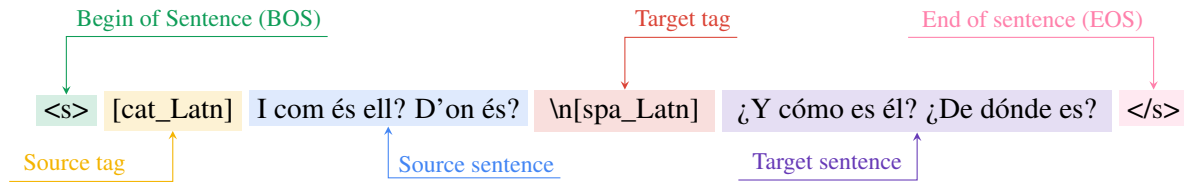

\subsection{Tokenizer} Prior studies have shown that vocabulary overlap plays a crucial role in zero-shot translation for encoder-decoder architectures \cite{stap-etal-2023-viewing, tan-monz-2023-towards}. More related to our work concerning tokenizer size in decoder-only architectures is the study by \citet{ali2023tokenizer}, who found that larger vocabulary sizes lead to improved downstream performance in multilingual settings. The main difference is that our focus is in Multilingual Neural Machine Translation (MNMT) while \citet{ali2023tokenizer} focused on more general multilingual tasks (Natural language inference, Question Answering, etc.). 

To investigate the impact of vocabulary sharing on zero-shot MNMT for decoder-only architectures, we train 3 tokenizers using BPE \cite{sennrich-etal-2016-neural} from the \textit{Huggingface tokenizer} library \cite{Moi_HuggingFace_s_Tokenizers_2023} with different vocabulary sizes; 32k, 128k, and 256k. Regarding the training data used to train the tokenizer, recent work has shown that while NMT performance is relatively robust to language imbalance, better performance is often achieved when languages are more equally represented in the training data \cite{zhang-etal-2022-robust}. In this work, we equally sample Romance languages and we oversample English, Basque, and German to avoid underrepresenting these languages and to achieve near parity \cite{petrov2024language} and fertility among all language pairs. More details about tokenizer experiments can be found in Appendix \ref{sec:tokenizer}.


\subsection{Model}

We trained one model for each of our three tokenizers using the same architecture as \textsc{Gemma}\ 2B\footnote{\url{https://huggingface.co/google/gemma-2b}} \cite{gemmateam2024gemma} to train a 2 billion parameter, transformer-based, decoder-only model. Following the scaling law proposed by \cite{NEURIPS2022_c1e2faff}, each model was trained on 30.9 billion words, corresponding to 54.7, 46.8, and 44.6 billion tokens for vocabularies of 32k, 128k, and 256k respectively. Details about the model size and model architecture can be found in Appendix \ref{sec:model_architecture}.


\subsection{Training}

We train all \parlam\ models with a context window of 2048 tokens, utilizing the Adam optimizer \cite{Kingma2014AdamAM} and the causal language modeling objective. Note that the main focus of this study is to understand how LLMs perform translation. Thus, \parlam\ models are not trained for state-of-the-art performance on MNMT. A more detailed description of the training configuration can be found in Appendix \ref{sec:model_training}.


\paragraph{Formatting} Figure \ref{sentence_example} presents an example of a formatted sentence for the Catalan to Spanish translation direction. During batching, we concatenate formatted sentences up to a context length of 2048 tokens, mixing different translation directions within a single batch. Padding is added to fill out the remainder of the sequence.


\subsection{Evaluation}

To compute reference-based translation quality we use \textsc{COMET-22} \cite{rei-etal-2022-comet} and \textsc{BLEU} \cite{papineni-etal-2002-bleu} metrics on the \flores\ devtest \cite{nllbteam2022language} and NTREX-101 \cite{federmann-etal-2022-ntrex} datasets. We additionally report \textsc{ChrF} \cite{popovic-2015-chrf} and \textsc{COMET-Kiwi-22} \cite{rei-etal-2022-cometkiwi} in appendix \ref{sec:detailed_results}. We use \textsc{TowerEval}\footnote{\textsc{TowerEval} uses the sacreBLEU implementation to compute \textsc{BLEU} and \textsc{ChrF} metrics.} \cite{alves2024tower} to compute all the evaluation metrics. For inference, we use beam search decoding with a beam size of 5 and limiting the translation length to 512 tokens.

We compare \parlam\ models with the following bilingual and multilingual models.

\begin{itemize}
    \item \nllb\ \cite{nllbteam2022language}: A transformer encoder-decoder model that supports 202 languages. We use the 600 million, the 1.3 billion, and the 3.3 billion parameter variants.

        \item Bilingual models \textsc{BSC}: Transformer encoder-decoder models, trained from scratch on language pairs that include Catalan. These models were developed as part of the Aina Project\footnote{\url{https://huggingface.co/projecte-aina}}. 
    
\end{itemize}

It is important to note that \nllb\ has seen parallel data for our zero-shot directions, therefore zero-shot only describes the condition in \parlam\ models. Our setup is designed to study the potential of a decoder-only architecture to perform zero-shot translation, specifically using Catalan as the pivot language.


\section{Results}

\begin{table*}[t]
    \large
    \centering
    \begin{tabular}{l@{\hskip 0.4in}cccccccc}
        \toprule
        & \multicolumn{4}{c}{\small Supervised directions} & \multicolumn{4}{c}{\small Zero-shot directions} \\ 
        \cmidrule(r){2-5} \cmidrule(r){6-9} 
        \textbf{} & \multicolumn{2}{c}{{\normalsize \flores }} & \multicolumn{2}{c}{{\normalsize \ntrex}} & \multicolumn{2}{c}{{\normalsize \flores }} & \multicolumn{2}{c}{{\normalsize \ntrex }} \\ 
        \cmidrule(r){2-3} \cmidrule(r){4-5} \cmidrule(r){6-7} \cmidrule(r){8-9} 
        & {\small \bleu } & {\small \comet}  & {\small \bleu} & {\small \comet}  & {\small \bleu}  & {\small \comet}  & {\small \bleu}  & {\small \comet}  \\ 
        \midrule
        \nllb-3.3B & 32.02 & 0.87 & 30.48 & 0.85 & 28.97 & 0.86 & 28.74 & 0.84 \\ 
        \nllb-1.3B & 31.02 & 0.86 &  29.68 & 0.85 &  28.48 & 0.86 & 28.37 & 0.84 \\ 
        \nllb-600M & 29.24 & 0.85 &  28.37 & 0.84 &  27.04 & 0.85 & 27.25 & 0.84 \\ 
        Bilinguals \textsc{BSC} & 31.93 & 0.86 & 29.77 & 0.84 & - & - & - & - \\
        \cdashlinelr{1-9}
        \parlam\ 32k &  30.44 & 0.86 & 28.46 & 0.84 & 23.25 & 0.83 & 23.03 & 0.80 \\ 
        \parlam\ 128k & 30.81 & 0.86 & 28.78 & 0.84 & 23.97 & 0.83 & 23.53 & 0.81 \\ 
        \parlam\ 256k & 30.72 & 0.86 & 28.87 & 0.84 & 24.42 & 0.84 & 23.81 & 0.81 \\ 
        \bottomrule
    \end{tabular}
    \caption{Averaged \bleu\; and \comet\; scores on supervised and zero-shot directions for \flores\; devtest and \ntrex. }
    \label{tab:results}
\end{table*}

Table \ref{tab:results} shows results for all \parlam\ models aggregated by supervised and zero-shot directions. The \parlam\ 32k, 128k and 256k variants perform equally well in supervised directions, achieving similar \bleu\ and \comet\ scores for both \ntrex\ and \flores\ datasets. In supervised directions, \parlam\ models demonstrate competitive performance, matching the \comet\ scores of the Bilingual \textsc{BSC} models and achieving scores comparable to the \nllb\ variants.


In zero-shot directions, the \parlam\ models exhibit a decline in performance compared to supervised directions. However, the decline is more pronounced in the \bleu\ scores than in the \comet\ scores, indicating that the overall quality remains relatively robust. Specifically, the \parlam\ 256k variant achieves a \comet\ score of 0.84 on the \flores\ dataset and 0.81 on the \ntrex\ dataset, which, although lower than its supervised performance, still demonstrates its zero-shot translation capabilities when training using only Catalan as the bridge language.

\paragraph{Larger vocabulary sizes improve zero-shot translation.} The results in Table \ref{tab:results} show that higher vocabulary sizes consistently yield better zero-shot capabilities. Specifically, the \parlam\ 256k variant outperforms the 32k and 128k variants in zero-shot scenarios for both \flores\ and \ntrex\ datasets.

To further understand the influence of the vocabulary size in zero-shot translation quality, we calculated the vocabulary overlap \cite{tan-monz-2023-towards} for each zero-shot translation direction as follows:

\begin{equation}
Overlap = \frac{| V_{src} \cap V_{tgt} | }{| V_{tgt} |}
\end{equation}

where \( V_{src}, V_{tgt}  \) are the set of unique words in the source and target language vocabulary respectively. We show the correlation between vocabulary overlap and both \textsc{BLEU} and \comet\ for zero-shot directions in table \ref{tab:vocab_overlap}. On average there is a positive correlation between the vocabulary overlap and the translation quality of 0.3 for \bleu\ and 0.57 for \comet, which diminishes as vocabulary size increases. This suggests that vocabulary overlap between the source and target languages further helps explain zero-shot performance, particularly for smaller vocabulary sizes.

\begin{table}[ht]
    \large
    \centering
    \begin{tabular}{lrrr}
    \toprule
       & {\tiny \parlam \; \small 32k}  & {\tiny \parlam \; \small 128k} & {\tiny \parlam \; \small 256k} \\
    \midrule
    {\normalsize \bleu} & 0.351  & 0.280 & 0.255 \\
    {\normalsize \comet} & 0.593 & 0.588 & 0.538 \\
    \bottomrule
    \end{tabular}
    \caption{Correlation between vocabulary overlap and \bleu, \textsc{COMET} metrics for different vocabulary sizes in zero-shot directions.}
    \label{tab:vocab_overlap}
\end{table}

\begin{figure*}[!t]
  \includegraphics[width=\linewidth]{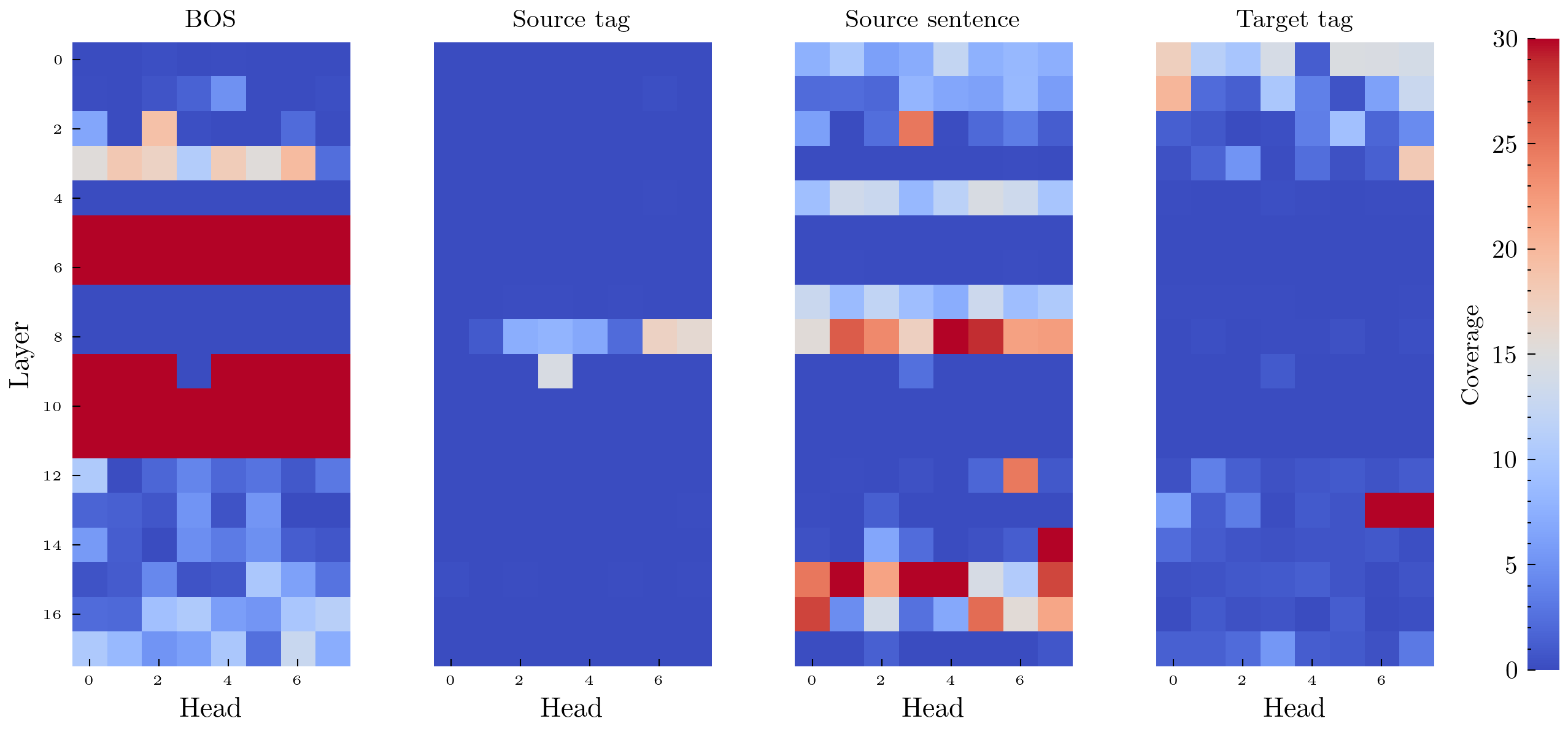}
  \caption {Coverage evaluating on \textsc{Flores-200} devtest using \parlam\ 32k. Each heatmap for each part of the prompt shows the coverage scores for each layer (vertical axis) and for each head (horizontal axis) in the model.  }\label{heatmapcoverage32}
\end{figure*}


\subsection{Understanding translation with an LLM}

Our goal is to understand how an LLM performs translation. We start by examining which parts of the prompt the model focuses on. This helps us determine the most important attention heads for each section of the prompt. Then, we study the model's cross-lingual representation space by extracting contextualized token embeddings.

\subsection{Attention}\label{prompt_head_importance}

For each attention head, we assess its importance by calculating coverage as defined by \cite{tu-etal-2016-modeling}. Originally, coverage was proposed for encoder-decoder attention and refers to the total attention a source token receives from target tokens. We adapt coverage for masked-self attention. Given a set of prompt's tokens $I$, the coverage formula for a single sentence is defined as:

\begin{equation}
\mathrm{cov}_{I}(\mathrm{head}) = \sum_{j \in J} \left(\sum_{i \in I} \alpha_{i,j}\right)^{2}
\end{equation}

where $\alpha_{i, j}$ denotes the attention weight from token $i$ to token $j$ and $J$ represent the set of the decoded (target) tokens.

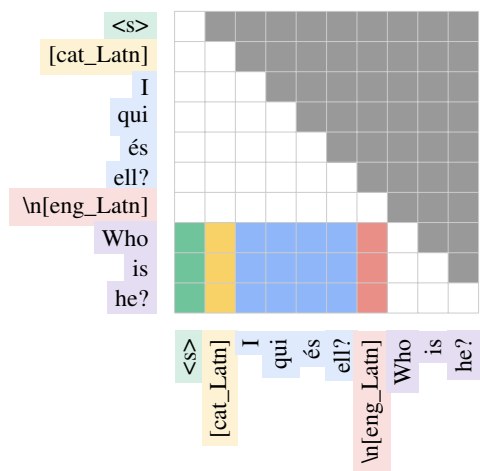
\begin{figure}[!b]
  \centering
     \begin{tikzpicture}
    
        \foreach \x/\y in {0.4/3.6, 0.8/3.2, 1.2/2.8, 1.6/2.4, 2.0/2.0, 2.4/1.6, 2.8/1.2, 3.2/0.8, 3.6/0.4} {
          \fill[black!40!white] (\x,\y) rectangle (4, \y+0.4);
        }

    
        \fill[custom1!60!white] (0.0,0.0) rectangle (0.4,1.2);
    
        \fill[custom2!60!white] (0.4,0.0) rectangle (0.8,1.2);
    
        \fill[custom3!60!white] (0.8,0.0) rectangle (2.4,1.2);
    
        \fill[custom4!60!white] (2.4,0.0) rectangle (2.8,1.2);
        
        \node[anchor=east] at (-0.1,3.8) {\footnotesize \highlight{custom1}{\text{<s>}} };
        \node[anchor=east] at (-0.1,3.4) {\footnotesize \highlight{custom2}{\text{[cat\_Latn]}} };
        \node[anchor=east] at (-0.1,3.0) {\footnotesize \highlight{custom3}{\text{I}} };
        \node[anchor=east] at (-0.1,2.6) {\footnotesize \highlight{custom3}{\text{qui}} };
        \node[anchor=east] at (-0.1,2.2) {\footnotesize \highlight{custom3}{\text{és}} };
        \node[anchor=east] at (-0.1,1.8) {\footnotesize \highlight{custom3}{\text{ell?}} };
        \node[anchor=east] at (-0.1,1.4) {\footnotesize \highlight{custom4}{\text{ \textbackslash n[eng\_Latn]}} };
        \node[anchor=east] at (-0.1,1.0) {\footnotesize \highlight{custom5}{\text{Who}} };
        \node[anchor=east] at (-0.1,0.6) {\footnotesize \highlight{custom5}{\text{is}} };
        \node[anchor=east] at (-0.1,0.2) {\footnotesize \highlight{custom5}{\text{he?}} };

        \node[anchor=east, rotate=90] at (0.2,-0.1) {\footnotesize \highlight{custom1}{\text{<s>}} };
        \node[anchor=east, rotate=90] at (0.6,-0.1) {\footnotesize \highlight{custom2}{\text{[cat\_Latn]}} };
        \node[anchor=east, rotate=90] at (1.0,-0.1) {\footnotesize \highlight{custom3}{\text{I}} };
        \node[anchor=east, rotate=90] at (1.4,-0.1) {\footnotesize \highlight{custom3}{\text{qui}} };
        \node[anchor=east, rotate=90] at (1.8,-0.1) {\footnotesize \highlight{custom3}{\text{és}} };
        \node[anchor=east, rotate=90] at (2.2,-0.1) {\footnotesize \highlight{custom3}{\text{ell?}} };
        \node[anchor=east, rotate=90] at (2.6,-0.1) {\footnotesize \highlight{custom4}{\text{ \textbackslash n[eng\_Latn]}} };
        \node[anchor=east, rotate=90] at (3.0,-0.1) {\footnotesize \highlight{custom5}{\text{Who}} };
        \node[anchor=east, rotate=90] at (3.4,-0.1) {\footnotesize \highlight{custom5}{\text{is}} };
        \node[anchor=east, rotate=90] at (3.8,-0.1) {\footnotesize \highlight{custom5}{\text{he?}} };

        \draw[step=0.4cm,gray!40!white,line width=0.001mm] (0,0) grid (4,4);
        
      \end{tikzpicture}
  \caption{Illustration of the regions in the attention matrix used to compute coverage for each part of the prompt.  We show the cross-attention regions between decoded tokens and the BOS, source tag, source sentence and target tag tokens in green, yellow, blue, and red, respectively. } 
  \label{fig:masked_sa}
\end{figure}

Each coverage metric is computed and averaged over the \flores\ devtest for each head in the model and for each translation direction. To understand which part of the prompt the model is focusing on in each head we study coverage separately for different parts of the prompt: BOS, source tag, source sentence and target tag. Figure \ref{fig:masked_sa} shows a graphical illustration of the regions in the attention matrix that are used to compute coverage based on the part of the prompt.


In Figure \ref{heatmapcoverage32}, we show the average coverage across all translation directions for each part of the prompt, employing \parlam\ 32k. We note that heads within the same layer generally exhibit similar coverage patterns. Future work may investigate how these patterns arise and how they are related to the usage of Multi-Query attention\footnote{When we use Multi-Query attention with \textit{num\_kv\_heads} set to 1,
the keys and values are shared across all heads from a specific layer and is only the query that differs which may hinder the specialization of the heads.} \cite{shazeer2019fast}. 

We find that source tag is the part of the prompt with least coverage. However, BOS, source sentence and target tag tokens exhibit varying degrees of coverage with some coverage spikes in specific layers and heads. Interestingly, layers 5, 6, 10 and 11 show coverage uniquely for the BOS token which suggests that all attention mass is given to the BOS token, leaving the residual stream unchanged. This patterns have recently been observed in auto-regressive language models and are named attention sink mechanisms \cite{xiao2023efficient, ferrando2024information, ferrando2024primer, cancedda2024spectral}. For instance, \citet{cancedda2024spectral} demonstrates that in Llama 2, the feed-forward blocks embed crucial information into the residual stream of the BOS token, enabling the attention sink mechanism to happen in subsequent layers. We show in appendix \ref{sec:coverageapp} the coverage heatmaps for \parlam\ 128k and 256k.



\paragraph{Source tag importance}\label{sec:source_tag_importance}

As previously pointed out, the source tag receives less attention than the other parts of the prompt. Specifically, it has an average coverage of 0.56 which is 3.7 times less coverage than the target token or 18.5 times less coverage than the BOS token. This motivates our next experiments which consist of evaluating \parlam\ models without indicating the source language. Specifically, we replace the source tag with another BOS token to mantain the same learned positional encodings and evaluate the model's performance on \flores\ devtest using BLEU. Table \ref{tab:removesource} shows the relative BLEU change with respect to the original model aggregated by language pair. Results show varying impacts across different language pairs when the source tag is omitted. For languages like English, French and Basque, the drop in \textsc{BLEU} scores is particularly significant. However, for other translation directions like Spanish and Catalan, the decrease in \textsc{BLEU} scores is negligible. This suggests that the model is more reliant on the source tag to represent certain languages, particularly those which are less related to the bridge language or those that the model has seen less during training.

Regarding the vocabulary size, the model with a 256k vocabulary shows the smallest average decrease in \textsc{BLEU} scores, suggesting that a larger vocabulary may improve the model's representation of the source language.

\begin{table}
  \centering
  \begin{tabular}{rrrr}
    \toprule
    & {\scriptsize \parlam \ \small 32k}  & {\scriptsize \parlam \ \small 128k} & {\scriptsize \parlam \ \small 256k} \\
    \midrule
    ca$\rightarrow$xx & -1.80 & -0.54  & -0.83 \\
    es$\rightarrow$xx & -0.43 & 0.23 & -0.33 \\
    pt$\rightarrow$xx & -8.13 & -6.01 & -5.54 \\
    gl$\rightarrow$xx & -6.52 & -4.18 & -4.92 \\
    it$\rightarrow$xx & -6.57 & -10.79 & -5.03	\\
    fr$\rightarrow$xx & -13.16	& -19.90 & -17.63 \\
    de$\rightarrow$xx & -7.54 & -2.73 & -6.73 \\
    en$\rightarrow$xx & -19.83 & -25.52	& -20.03 \\
    eu$\rightarrow$xx & -16.73	& -11.03 & -13.23 \\
    \cdashlinelr{1-4}
    \textbf{Avg.} & -8.97 & -8.94 & -8.25 \\
    \bottomrule
  \end{tabular}
  \caption{Relative \textsc{BLEU} change with respect to \parlam\ models after ignoring the source tag. We label languages according to their BCP-47 language code (see Table \ref{tab:bcp} from Appendix \ref{sec:dataset}). }
  \label{tab:removesource}
\end{table}

\paragraph{Redundant heads} Previous work on MNMT has shown that coverage is a good indicator for pruning cross attention heads in encoder-decoder architectures \cite{kim-etal-2021-multilingual}. Following \citet{kim-etal-2021-multilingual}, we study whether coverage can be used to prune heads in a decoder-only architecture without sacrificing the model's performance. Specifically, we mask all attention heads within a specific layer that fall below a predetermined coverage threshold. We compute coverage per layer for a specific direction as follows:

\begin{equation}\label{eq:coverage_layer}
\begin{aligned}
    \text{COV}_l = \phi (\sum_{i=1}^{H} \sum_{j \in \text{Pr}} 
    \mathrm{cov}_j\left(\text{head}_{l,i}\right) ) \\
    \text{ \footnotesize{\text{Pr} = \{\text{BOS}, \text{Source tag}, \text{Source sentence}, \text{Target tag}\}} }
\end{aligned}
\end{equation}

where \( \text{COV}_l \) represents the coverage of layer \( l \), \( H \) is the total number of attention heads in the model, and \( \mathrm{Pr} \) is a set that contains sets of tokens for each part of the prompt. Finally, $\phi$ is a MinMax Scaler used to normalize the metric between 0 and 1.

\begin{figure*}[!t]
  \includegraphics[width=\linewidth]{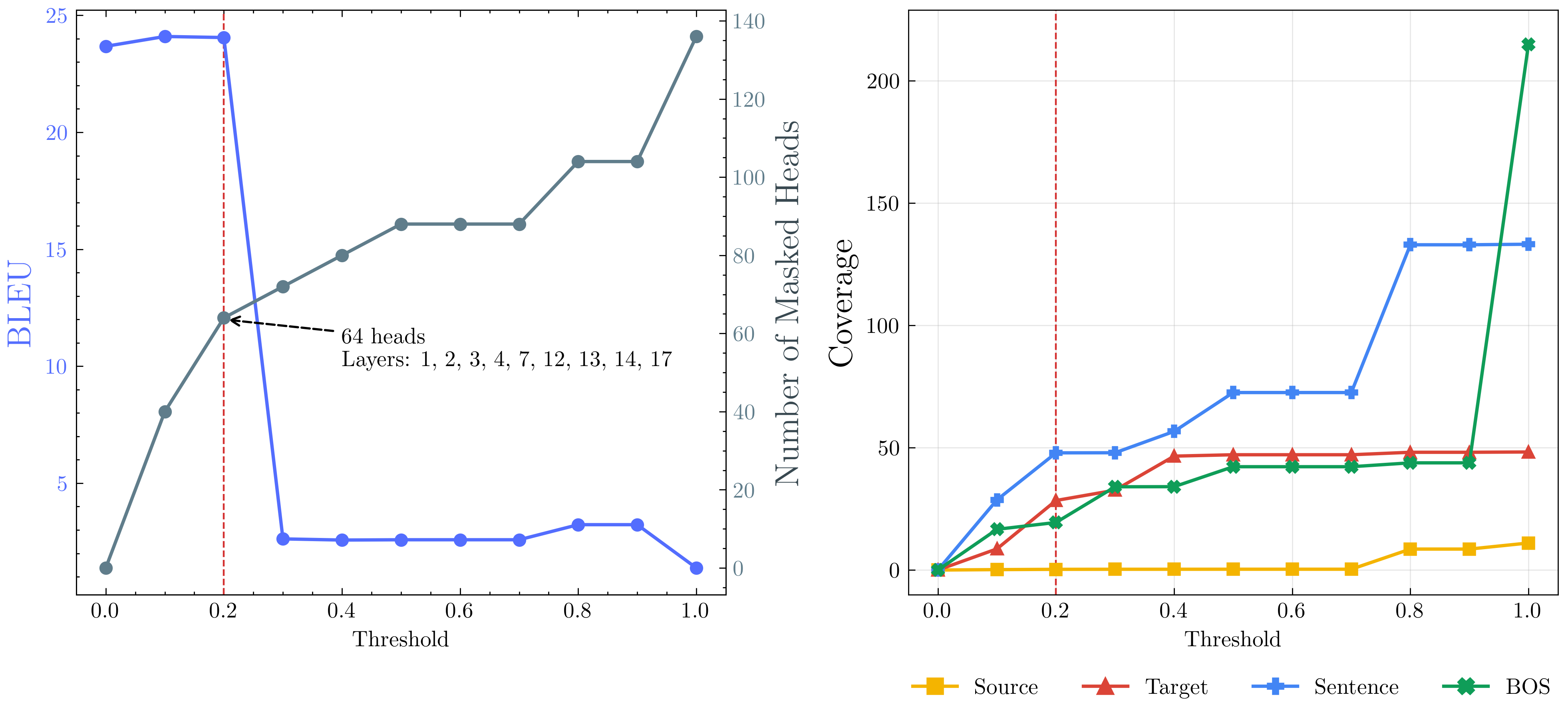}
  \caption{Impact of masking on BLEU score and number of masked heads across different coverage thresholds (left). Accumulated coverage of masked heads for source tag, target tag, source sentence, and BOS (right). Experiments are evaluated on the Spanish to Catalan direction.}
  \label{fig:prunning_plot}
\end{figure*}

We use \flores\ devtest to evaluate the impact of masking heads per layer based on the coverage criterion (Equation \ref{eq:coverage_layer}). Figure \ref{fig:prunning_plot} (left) illustrates the evolution of \textsc{BLEU} scores as we mask heads in \parlam\ 32k for the Spanish to Catalan direction (supervised). The right axis indicates the number of heads that are masked. We find that up to 64 heads can be masked without degrading the model's performance using a threshold of 0.2, representing 47.05\% of the model's total heads. In Figure \ref{fig:prunning_plot} (right), we show the cumulative coverage for the different parts of the prompt. We observe that for a threshold of 0.2, the masked heads represent 9.05\%, 2.61\%, 36\% and 58.9\% total coverage for the BOS, source tag, source sentence and target tag tokens respectively. This indicates that the majority of the masked heads are paying attention to the target tag token and to a lesser extent to the source sentence tokens. This suggests that these heads are less critical for maintaining translation quality. Specifically, when masking these 64 heads we are only using heads from layers 5, 6, 8, 9, 10, 11, 15, and 16 which are the layers with higher coverage for the BOS, source tag and source sentence tokens (see Figure \ref{heatmapcoverage32}). Regarding the source tag, we find that even though it is the part of the prompt with the lowest coverage, it is still useful for maintaining the translation quality. This observation aligns with our previous findings from section \ref{sec:source_tag_importance}.

In table \ref{tab:pruneheads} from appendix \ref{sec:redundant_heads} we report the number of heads that we can mask without losing more than 2 \textsc{BLEU} points for different translation directions and for different vocabulary sizes. We find that for larger vocabulary sizes we can mask a higher number of heads without hurting the model's performance. Specifically, with \parlam\ 256k, we can mask 88 heads on average, which represents 64.7\% of the total number of heads. Future work may investigate how this can be used for model pruning.

\subsection{Language subspaces}

\paragraph{Subspace distances} We first extract sub-word tokens output by each layer in the Transformer. Specifically, we use the first 300 sentences from \flores\ devtest for each source language, denoted as $s$. These sentences are used to create translation prompts from $s$ to each target language (300 * 8 = 2,400 prompts). For each prompt, we extract the token embeddings from each layer of the model and concatenate the consecutive tokens to form $\mathbf{H}_l^s$. Then, we apply singular value decomposition (SVD) on $\mathbf{H}_l^s$ after substracting the mean. We calculate pairwise distances among the 9 languages using the affine subspace for each language computed by the SVD, utilizing the Riemannian metric on the space of positive definite matrices described in \cite{chang-etal-2022-geometry}, which is both symmetric and invariant to affine transformations.

\begin{figure}[t]
  \includegraphics[width=\columnwidth]{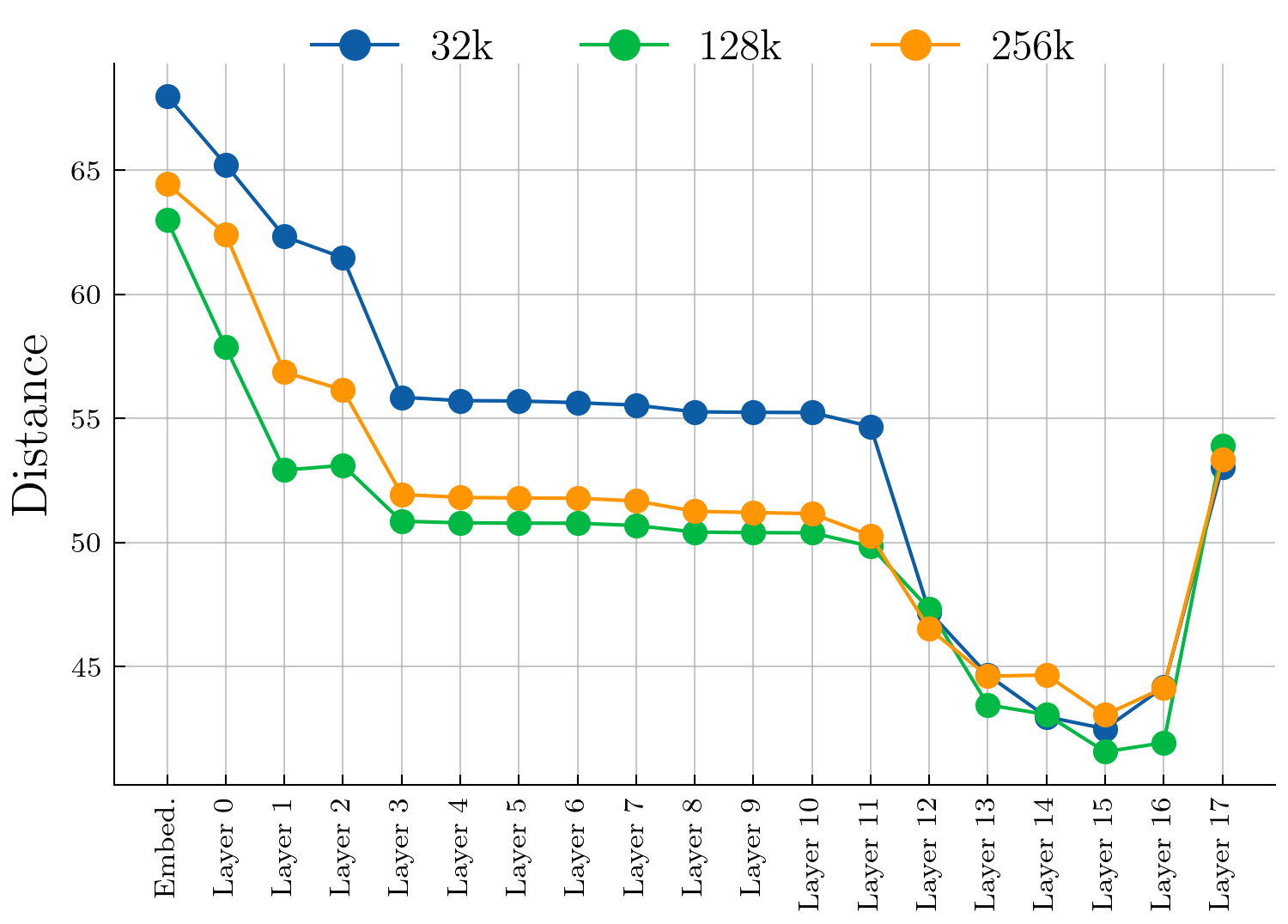}
  \caption{Mean distance between language subspaces grouped by vocabulary size. Additional plots grouped by languages and vocabulary sizes are included in Appendix \ref{sec:subspace_distances}.}
  \label{fig:distances_mean}
\end{figure}

\begin{figure*}[t]
  \includegraphics[width=\linewidth]{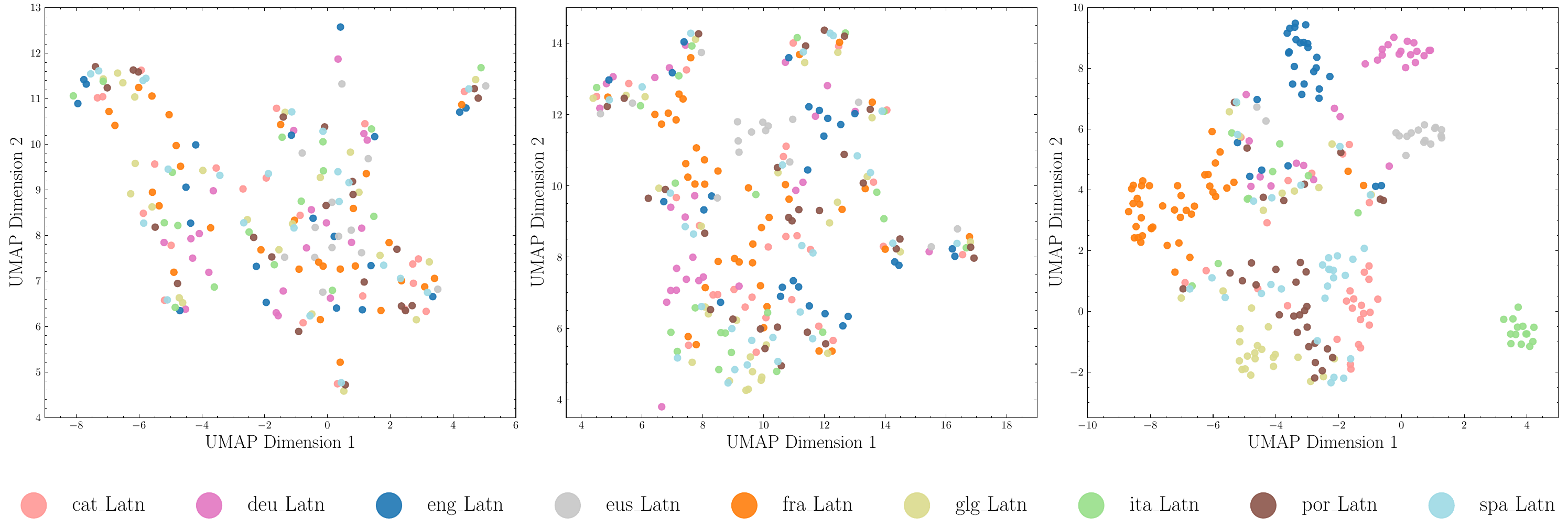}
  \caption{From left to right: Representations at the token embeddings in the embedding layer, the penultimate layer, and the last layer of \parlam\ 32k grouped by source language. See Appendix \ref{sec:umap_plots_appendix} for more additional plots.}\label{fig:umap_visualization}
\end{figure*}

Figure \ref{fig:distances_mean} shows the mean distance between language subspaces in each layer. As we can see, the distance between language subspaces decreases with model depth. Initially, from the embeddings layer to layer 0 we can observe a significant decrease of approximately 5.07\%, and from layer 0 to layer 1, a further reduction of 7.23\%. In middle layers (layers 3 to 11), distances are relatively stable and show minimal variations. This suggests that the model applies only minimal transformations to the representations along these layers. Interestingly, in layer 8 we can observe a small decrease in the distance of 0.05\% which we hypothesize may be due to the model's attention focusing more on the source token at this layer (see Figure \ref{heatmapcoverage32}). As we move to the deeper layers, the distances continue to decrease, with a significant drop of approximately 8.88\% from layer 11 to layer 12, a trend that persists through layers 12 to 16. However, in the last layer, there is a notable increase in distance by approximately 23.06\%. These results align with previous work on encoder-only models, which suggest that in intermediate layers the model representations diverge more from the embedding layer representation and from the final layer. Both the embedding layer and the final layer are highly language-sensitive \cite{chang-etal-2022-geometry, libovicky-etal-2020-language, pires-etal-2019-multilingual}.

Regarding the vocabulary size, as shown in Figure \ref{fig:distances_mean}, we observe that for \parlam\ 32k the distance between embeddings are higher than \parlam\ 128k or \parlam\ 256k until layer 12, where distances become similar. This can be attributed to the higher vocabulary overlap between languages in \parlam\ 32k, where each embedding represents a more diverse concept, limiting its ability to learn language-agnostic representations. 

\paragraph{Visualization} In the previous subsection, we found that the distances between embeddings initially decrease, 
then in the last layer, the embeddings spread out. To understand this phenomenon, we visualize the token embeddings using Uniform Manifold Approximation and Projection (UMAP) \cite{mcinnes2018umap-software}. We construct prompts from each source language to Galician. Token embeddings per layer are concatenated to form $\mathbf{P}_l^s$, then we apply UMAP to reduce the dimensionality of the representations.

Figure \ref{fig:umap_visualization} shows the UMAP visualizations for token embeddings in the embedding layer and the two last layers of the model coloured by source language. As we can see, token embeddings remain language-neutral as they pass through the model until the last layer, where token embeddings group by source language. This suggests that the model must align embeddings cross-linguistically until reaching the last layer where it clusters by source language. This explains the distance of the last layer (see Figure \ref{fig:distances_mean}). See Appendix \ref{sec:umap_plots_appendix} for more additional plots\footnote{Additionally, we include UMAP Spherical Voronoi diagrams as supplementary materials (see Appendix \ref{spherical_voronoi_app}).}.


\section{Conclusions}

This work demonstrates the successful training of an LLM-based machine translation system from scratch using only parallel data. The achieved results are comparable to those of existing encoder-decoder architectures for supervised translation tasks. We identified that larger vocabulary sizes consistently improve translation quality across zero-shot directions, suggesting the potential benefits of experimenting with even larger or language-specific vocabularies.

Further analysis revealed that different LLM layers focus on distinct aspects of the prompt, particularly the source language tag, which exhibits significant language variation. By leveraging this insight and employing an appropriate criterion, we achieved a performance reduction of less than 2 BLEU score while removing over 47\% of attention heads. Additionally, our exploration of the learned cross-lingual space demonstrates that languages get closer in the cross-lingual space as they get to deeper layers and highlight the layers with the most significant impact on the learned space.

This research opens doors for further investigation. We identified "sink heads" that primarily focus on the BOS token. Exploring their utility and relationship to the learned cross-lingual representations presents an opportunity for future work. Additionally, further research into the optimization of vocabulary size along model size could also lead to better NMT models.


\section{Limitations}

This study focused on understanding the capabilities of an LLM trained solely on parallel data, without aiming to achieve state-of-the-art translation quality or extensive language support. Here are some key limitations to consider when interpreting the results:

\paragraph{Data Scope:} The experiment employed non-English centric data with a focus on Western, Latin-script languages. This approach aimed to isolate the impact of vocabulary size and overlap, but limits generalizability to languages with different scripts or historical connections. However, the inclusion of Basque, a non-Indo-European Subject-Object-Verb (SOV) language, provides valuable insights into the model's handling of structural variations.

\paragraph{Scalability:} The study did not explore the impact of model scale and data availability on translation across diverse languages and scripts. Further research is necessary to understand how these factors influence performance in more complex settings.

These two main aspects will be considered as future work by studying the scalability of these architectures on both model size and translation directions.

\section{Acknowledgements}

This work has been promoted and financed by the Government of Catalonia through the Aina project, by the Ministerio para la Transformación Digital y de la Función Pública and Plan de Recuperación, Transformación y Resiliencia - Funded by EU – NextGenerationEU within the framework of the project ILENIA with reference 2022/TL22/00215337, 2022/TL22/00215336, 2022/TL22/00215335, 2022/TL22/00215334, as well as by DeepR3 (TED2021-130295B-C32) founded by MCIN/AEI/10.13039/501100011033 and European Union NextGeneration EU/PRTR.

\bibliography{latex/acl_latex}

\appendix

\onecolumn

\section{Dataset}
\label{sec:dataset}

Table \ref{tab:data} shows the number of sentences and number of words per language pair in the created Catalan-Centric dataset.

\begin{table}[H]
    \begin{minipage}{0.45\textwidth}
        \begin{table}[H]
    \centering
    \begin{tabular}{rrr}
        \toprule
        \textbf{Pair} & \textbf{N sentences} & \textbf{N words} \\
        \cdashlinelr{1-3}
        ca {\tiny \textsc{SYN}} {\footnotesize $\leftrightarrow$} de & 187,483,456 & 6,847,140,698 \\
        ca {\footnotesize $\leftrightarrow$} de & 12,516,544 & 603,121,312 \\
        \cdashlinelr{1-3}
        ca {\tiny \textsc{SYN}} {\footnotesize $\leftrightarrow$} it & 181,034,146 & 6,526,304,128 \\
        ca  {\footnotesize $\leftrightarrow$} it & 18,965,862 & 577,243,404 \\
        \cdashlinelr{1-3}
        ca  {\footnotesize $\leftrightarrow$} es & 171,907,026 & 8,252,262,032 \\
        \cdashlinelr{1-3}
        ca {\tiny \textsc{SYN}} {\footnotesize $\leftrightarrow$} pt & 62,858,532 & 2,429,548,286 \\
        ca  {\footnotesize $\leftrightarrow$} pt & 12,319,262 & 504,959,082 \\
        \cdashlinelr{1-3}
        ca {\footnotesize $\leftrightarrow$} en & 60,046,068 & 2,429,961,320 \\
        \cdashlinelr{1-3}
        ca  {\footnotesize $\leftrightarrow$} fr & 37,269,716 & 1,114,635,790 \\
        \cdashlinelr{1-3}
        ca {\tiny \textsc{SYN}} {\footnotesize $\leftrightarrow$} eu & 17,998,782 & 749,042,034 \\
        ca  {\footnotesize $\leftrightarrow$} eu & 2,091,356 & 61,237,122 \\
        \cdashlinelr{1-3}
        ca {\tiny \textsc{SYN}} {\footnotesize $\leftrightarrow$} gl & 11,434,180 & 531,773,730 \\
        ca  {\footnotesize $\leftrightarrow$} gl & 7,713,022 & 263,280,596 \\
        \cdashlinelr{1-3}
        \textbf{Total} &  {783,637,952} & {30,890,509,534} \\
        \bottomrule
    \end{tabular}
    \caption{Number of sentences and words for each language pair. We label languages with their BCP-47 language code. \textsc{SYN} means synthetic data generated on the source side for the ca-xx direction. }
    \label{tab:data}
\end{table}
    \end{minipage}
    \hfill
    \begin{minipage}{0.3\textwidth}
        \begin{table}[H]
    \centering
    \begin{tabular}{l}
        \toprule
        \textbf{Dataset}  \\
        \cdashlinelr{1-1}
        Aina-ca-en-Parallel-Corpus \\
        CCAligned \\ 
        Covost2 \\ 
        DOGC \\
        EUBookshop \\ 
        Europarl \\ 
        Globalvoices \\ 
        Gnome \\ 
        HLPT \\ 
        KDE4 \\ 
        MultiCCAligned \\ 
        NLLB \\ 
        OpenSubtitles \\ 
        ParaCrawl \\ 
        Tatoeba \\ 
        TildeModel \\
        Ubuntu \\ 
        Wikimatrix \\ 
        Wikimedia \\ 
        XLEnt \\ 
        \bottomrule
    \end{tabular}
    \caption{Data sources.}
    \label{tab:datasources}
\end{table}
    \end{minipage}
    \hfill
    \begin{minipage}{0.2\textwidth}
        \vspace{2.78cm}
        \begin{table}[H]
    \centering
    \begin{tabular}{lr}
        \toprule
        \textbf{Language} & \textbf{Id} \\
        \cdashlinelr{1-2}
        Catalan & ca \\
        German & de \\
        English & en \\
        Spanish & es \\
        Basque & eu \\
        Italian & it \\
        Galician & gl \\
        French & fr \\
        Portuguese & pt \\
        \bottomrule
    \end{tabular}
    \caption{List of BCP-47 language codes.}
    \label{tab:bcp}
\end{table}
    \end{minipage}
\end{table}

\section{Tokenizer}
\label{sec:tokenizer}

In our experiments, we utilized the BPE algorithm \cite{sennrich-etal-2016-neural}  from the \textit{Huggingface Tokenizer} library \cite{Moi_HuggingFace_s_Tokenizers_2023}. The settings used for training the tokenizer are detailed in Table \ref{tab:tokenizer_config}. Every language tag is represented by a BCP-47 tag sequence where the base subtag is a three-letter ISO 639-3 code, followed by ISO 15924 script subtags.

\begin{table}[H]
\centering
\begin{tabular}{lr}
\toprule
\textbf{Hyper-Parameter}       & \textbf{Value(s)}   \\ 
\cdashlinelr{1-2}
model\_type               & BPE             \\
vocab\_size & 32k \& 128k \& 256k \\
nfkd\_normalizer & True \\
lowercase\_normalizer & False \\
pre\_tokenizer & ByteLevel \\
add\_prefix\_space & False \\
special\_tokens & <s>, </s>, <pad>, <mask>, 
 [deu\_Latn], \\
 &   [eng\_Latn], [eus\_Latn], [fra\_Latn], [glg\_Latn], \\
 &   [ita\_Latn], [por\_Latn], [spa\_Latn], [cat\_Latn] \\
\bottomrule
\end{tabular}
\caption{BPE tokenizer configuration.}
\label{tab:tokenizer_config}
\end{table}

We trained various tokenizers employing two distinct sampling strategies for each vocabulary size, then we  evaluated them on fertility and parity \cite{petrov2024language} metrics on \flores\ devtest. For a given tokenizer T and a set of sentences S, fertility is determined by dividing the total number of tokens generated from S (using T) by the total number of words in S. Parity is defined as achieving a balanced tokenization ratio between two languages. Specifically, a tokenizer T achieves parity for language \( A \) with respect to language \( B \) if the ratio \( \frac{|T(s_A)|}{|T(s_B)|} \approx 1 \), where \( s_A \) and \( s_B \) denote the sets of all sentences for languages \( A \) and \( B \), respectively. 

\begin{figure}[H]
  \includegraphics[width=\linewidth]{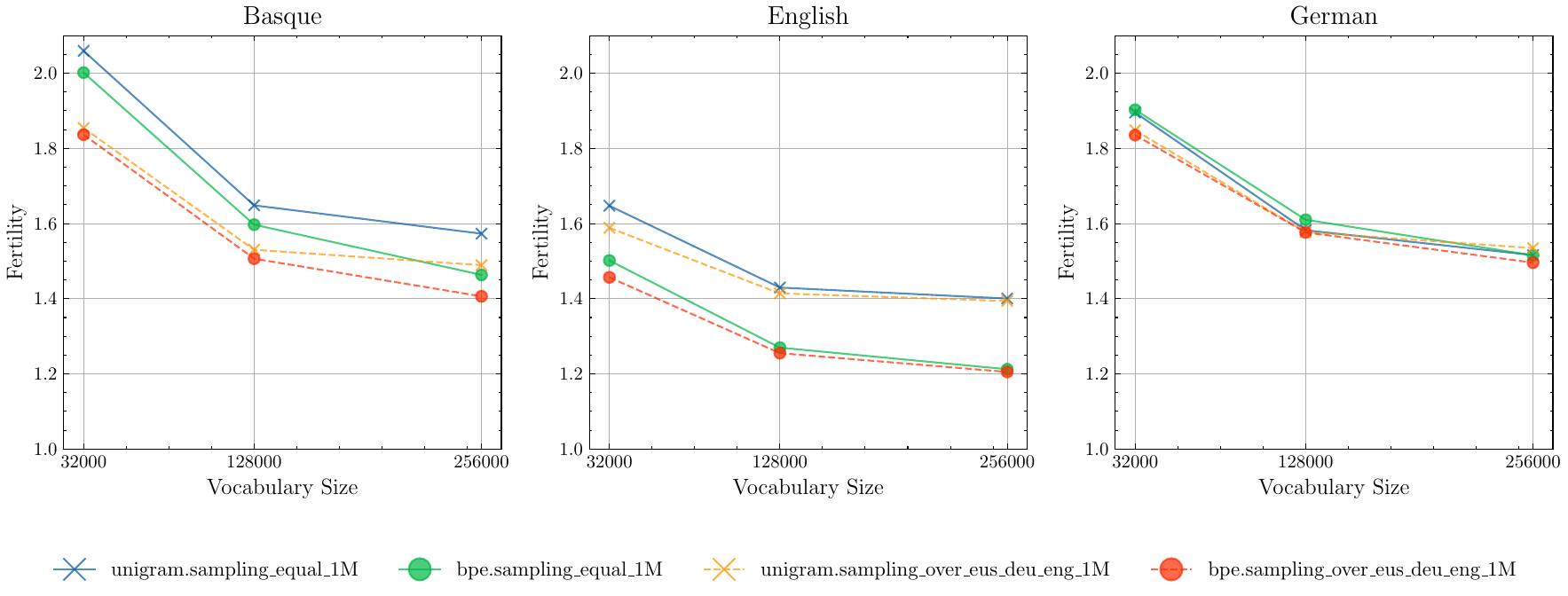}
  \caption {From left to right: fertility evaluated on Basque, English and German. Fertility is in the vertical axis, and vocabulary size is in the horizontal axis. }\label{tokenizationexperiments}
\end{figure}

\begin{figure}[H]
  \includegraphics[width=0.32\linewidth]{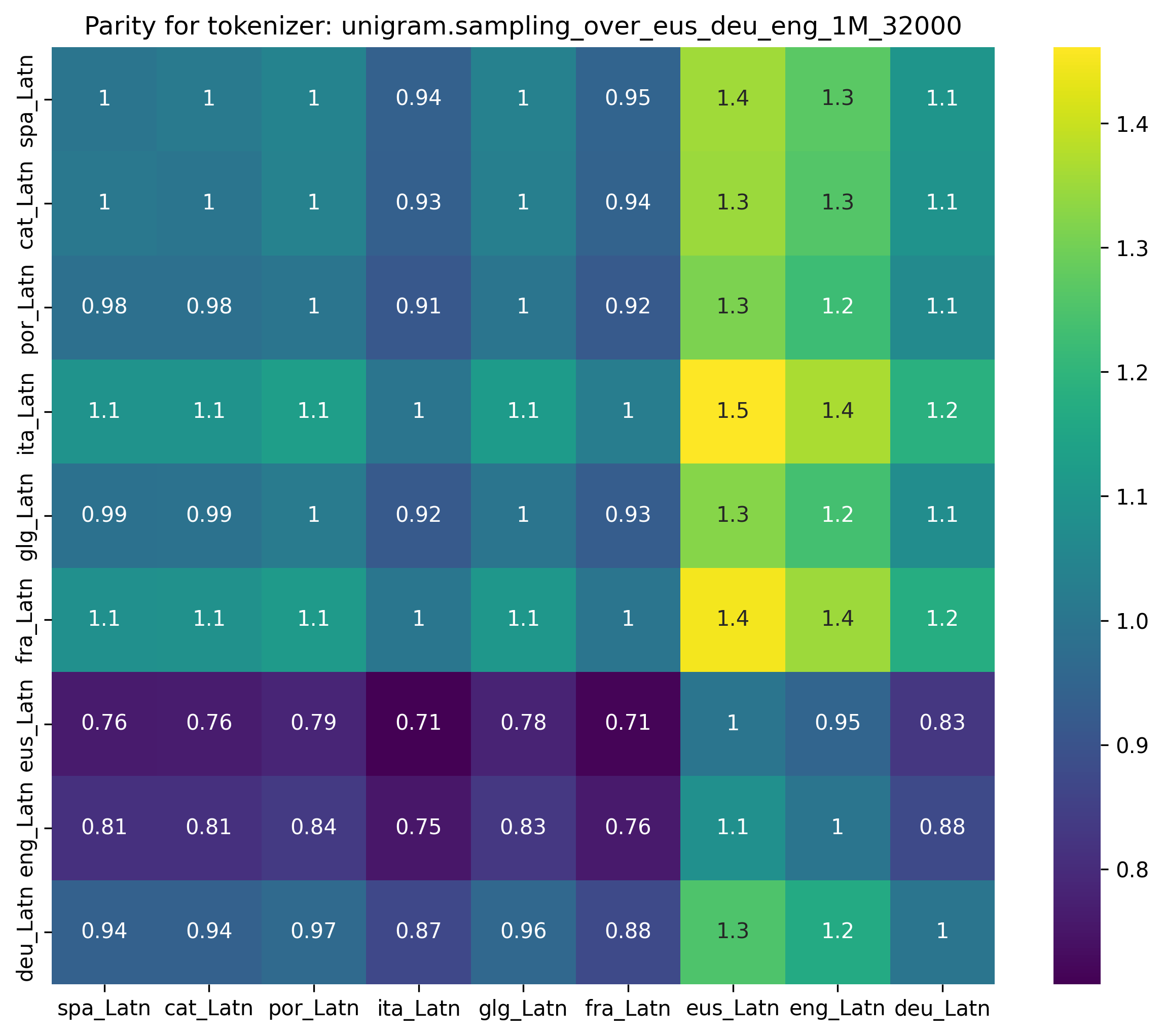} \hfill
  \includegraphics[width=0.32\linewidth]{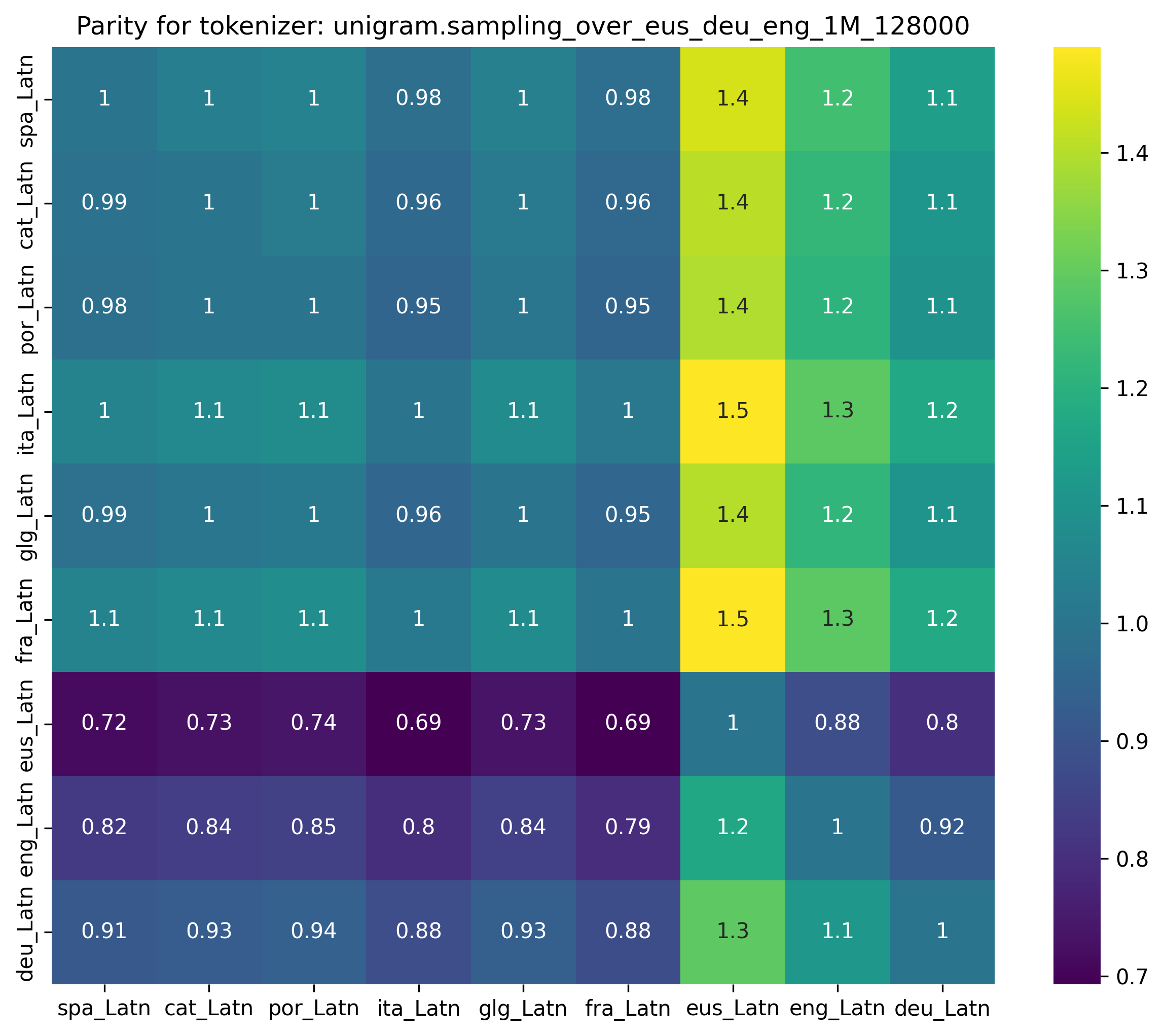}
  \includegraphics[width=0.32\linewidth]{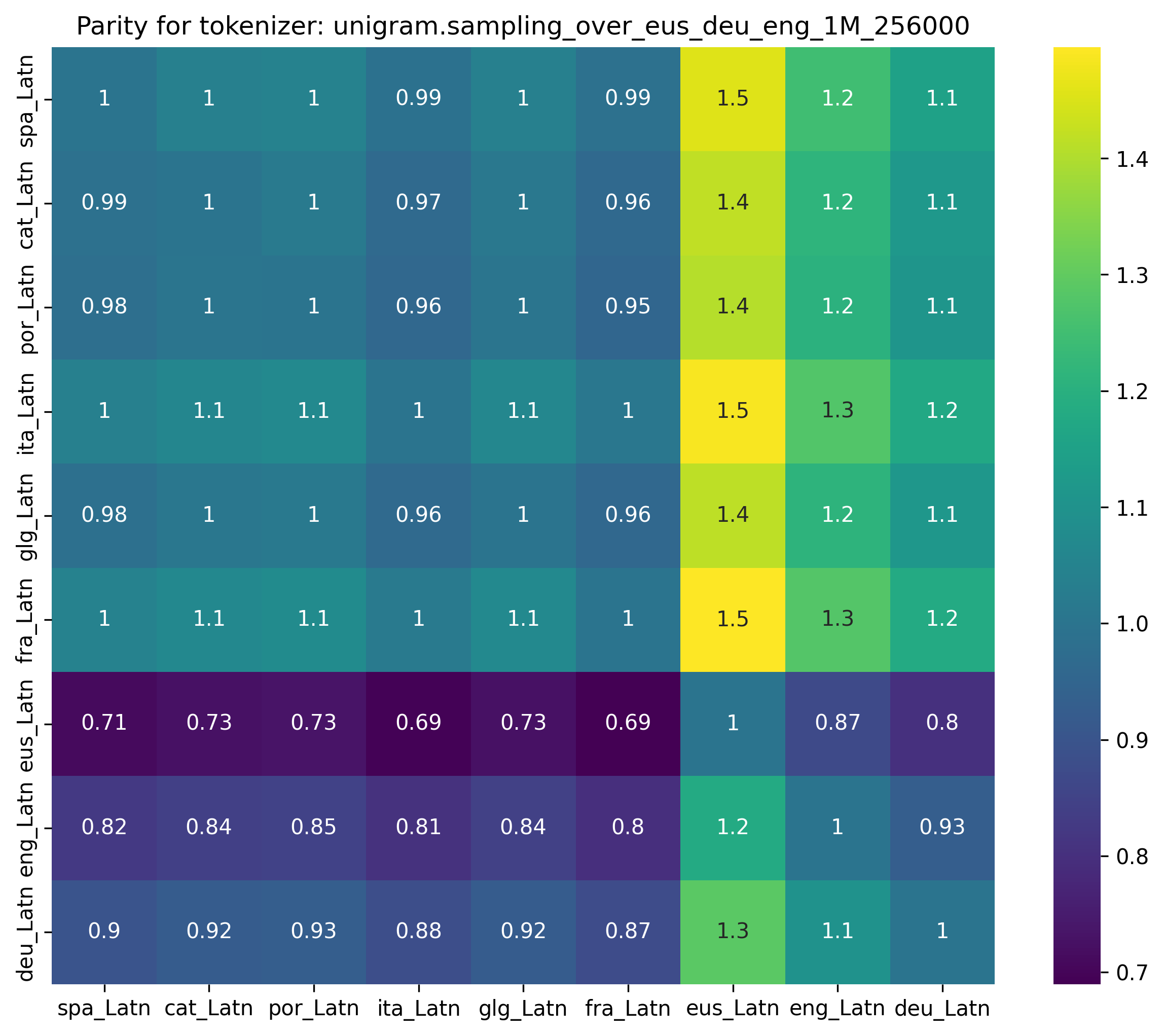}
  \caption {Parity for the different vocabulary sizes.}\label{fig:parity_tokenizers}
\end{figure}

We experimented with both unigram and BPE implementations from the \textit{Huggingface Tokenizer} library. We tested two sampling strategies: one involving the sampling of 1 million sentences from all languages, and another involving the equal sampling of 1 million sentences from Romance languages, with an oversampling of 3 million sentences for English, Basque, and German. Figure \ref{tokenizationexperiments} presents the fertility metrics on English, Basque, and German. Given the results, we decided to use the BPE algorithm with the oversampling strategy for our final experiments. We also report obtained parity metrics by vocabulary size in figure \ref{fig:parity_tokenizers} and average fertility (average of fertility per each language) per vocabulary size as well as the number of tokens in the dataset in Table \ref{tab:fertility} \footnote{We compute the number of tokens as Average Fertility * Number of words in the dataset. The number of words is 30,890,509,534.}.

\begin{table}[H]
\centering
\begin{tabular}{lrr}
\toprule
\textbf{Vocabulary size}       & \textbf{Average Fertility} & \textbf{Number of tokens}  \\ 
\cdashlinelr{1-3}
\parlam\ 32k               &  1.77     & 54.7B        \\
\parlam\ 128k               & 1.52     & 46.8B        \\
\parlam\ 256k               & 1.44     & 44.6B        \\

\bottomrule
\end{tabular}
\caption{Fertility and number of tokens in the dataset grouped by vocabulary size.}
\label{tab:fertility}
\end{table}

\section{Model Architecture}
\label{sec:model_architecture}

Table \ref{tab:hyperparameters} summarizes the architecture used for \parlam\ models.

\begin{table}[H]
\centering
\begin{tabular}{lr}
\toprule
\textbf{Hyper-Parameter}       & \textbf{Value}   \\ 
\cdashlinelr{1-2}
Hidden Dimension               & 2048             \\
Layers                         & 18               \\
Intermediate Size (in MLPs)        &       16384          \\
Attention-Heads                & 8               \\ 
Head size & 256 \\
Num KV Heads & 1 \\
Max Seq Length & 2048 \\
Position Embeddings            &    Rotary       \\ 
Rope Theta &  10000 \\
Precision   &  \texttt{float-32} \\ 
RMSNorm $\epsilon$ &  1e-06 \\
\bottomrule
\end{tabular}
\caption{Model architecture.}
\label{tab:hyperparameters}
\end{table}

\section{Training}
\label{sec:model_training}

For training, the learning rate is warmed up from $1 \times 10^{-7}$ to a maximum of $3 \times 10^{-4}$ over the first 2000 steps. We apply a weight decay of 0.1 and a gradient clipping of 1.0. During training, we set an effective batch size of 81,920 tokens per gradient step distributed over 40 NVIDIA H100-64GB GPUs. We use DeepSpeed with full \texttt{float32} training. 

\begin{table}[H]
\centering
\begin{tabular}{lr}
\toprule
\textbf{Hyper-Parameter}       & \\
\cdashlinelr{1-2}
Batch size                     & 40                 \\
Number of Epochs               & 1                 \\
Optimizer                      & Adam   \\ 
Adam-$\beta_1$                 &  0.9             \\ 
Adam-$\beta_2$                 &  0.999             \\ 
Adam-$\epsilon$                &  1e-08             \\ 
Learning rate                  &  3e-04          \\ 
LR Scheduler                &   Linear         \\ 
Warmup Steps                &   2000         \\ 
\bottomrule
\end{tabular}
\caption{Model training hyper-parameters}
\label{tab:training}
\end{table}

\begin{table}[H]
    \centering
    \begin{tabular}{ll}
        \toprule
        Num examples & 26,301,993 \\  
        Num tokens = Num examples * 2048 (considering pad tokens) & 53,866,481,664 \\  
        Num Epochs & 1 \\  
        Instantaneous batch size per device & 1 \\  
        Total train batch size (w. parallel, distributed \& accumulation) & 40 \\  
        Gradient Accumulation steps & 1 \\  
        Total optimization steps & 657,550 \\  
        Number of trainable parameters & 2,047,420,416 \\  
        \bottomrule
    \end{tabular}
    \caption{Training and performance information for \parlam\; 32k. }
    \label{tab:training_info32}
\end{table}
\begin{table}[H]
    \centering
    \begin{tabular}{ll}
        \toprule
        Num examples & 23,093,719 \\  
        Num tokens = Num examples * 2048 (considering pad tokens) & 47,295,936,512 \\  
        Num Epochs & 1 \\  
        Instantaneous batch size per device & 1 \\  
        Total train batch size (w. parallel, distributed \& accumulation) & 40 \\  
        Gradient Accumulation steps & 1 \\  
        Total optimization steps & 577,343 \\  
        Number of trainable parameters & 2,244,028,416 \\  
        \bottomrule
    \end{tabular}
    \caption{Training and performance information for \parlam\; 128k. }
    \label{tab:training_info128}
\end{table}
\begin{table}[H]
    \centering
    \begin{tabular}{ll}
        \toprule
        Num examples & 22,213,825 \\  
        Num tokens = Num examples * 2048 (considering pad tokens) & 45,493,913,600 \\  
        Num Epochs & 1 \\  
        Instantaneous batch size per device & 1 \\  
        Total train batch size (w. parallel, distributed \& accumulation) & 40 \\  
        Gradient Accumulation steps & 1 \\  
        Total optimization steps & 555,346 \\  
        Number of trainable parameters & 2,506,172,416 \\ 
        \bottomrule
    \end{tabular}
    \caption{Training and performance information for \parlam\; 256k. }
    \label{tab:training_info256}
\end{table}

\begin{figure}[H]
  \includegraphics[width=\linewidth]{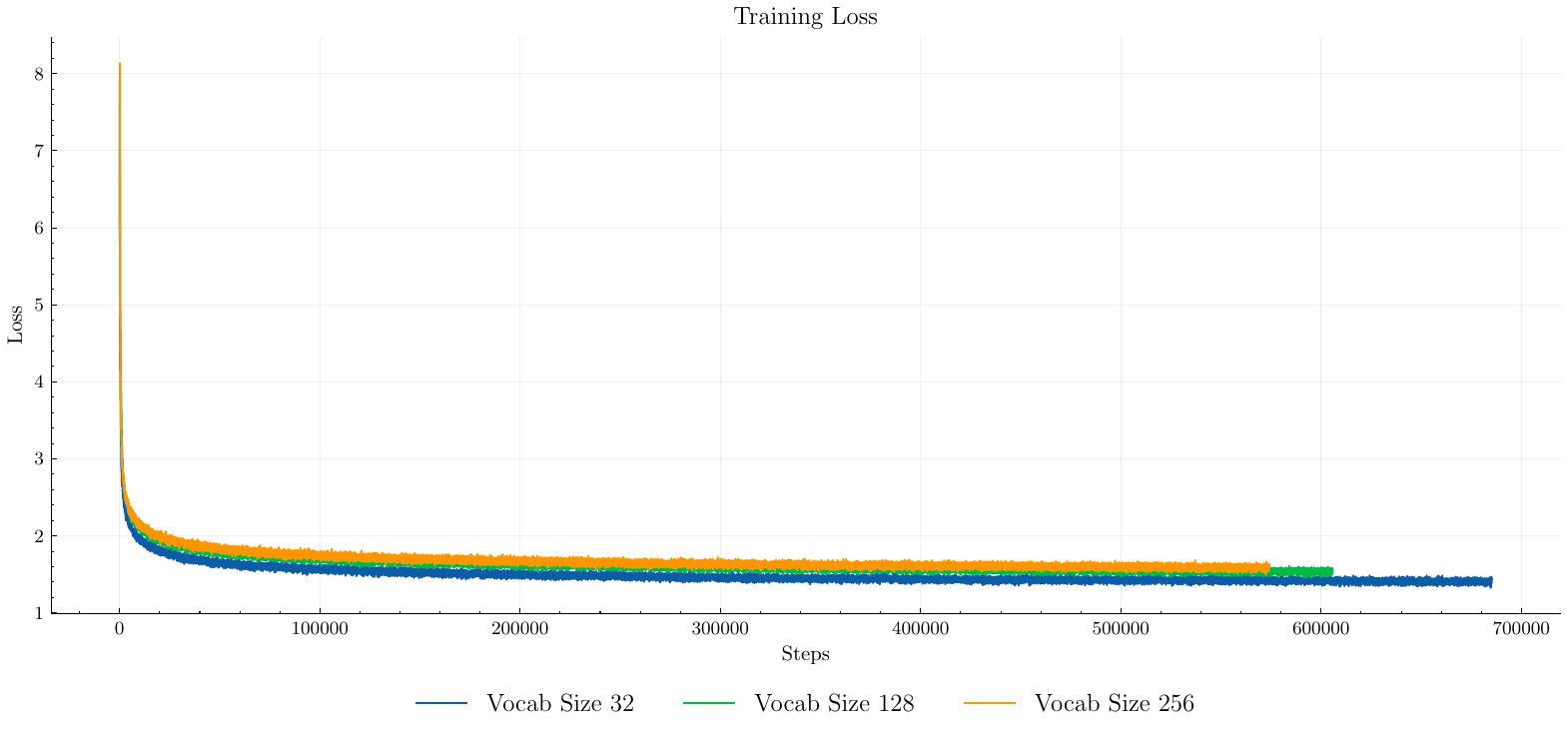}
  \caption {Training loss.  }\label{trainingloss}
\end{figure}

\section{Coverage metrics}\label{sec:coverageapp}

\begin{figure}[H]
  \centering
  \includegraphics[width=0.84\linewidth]{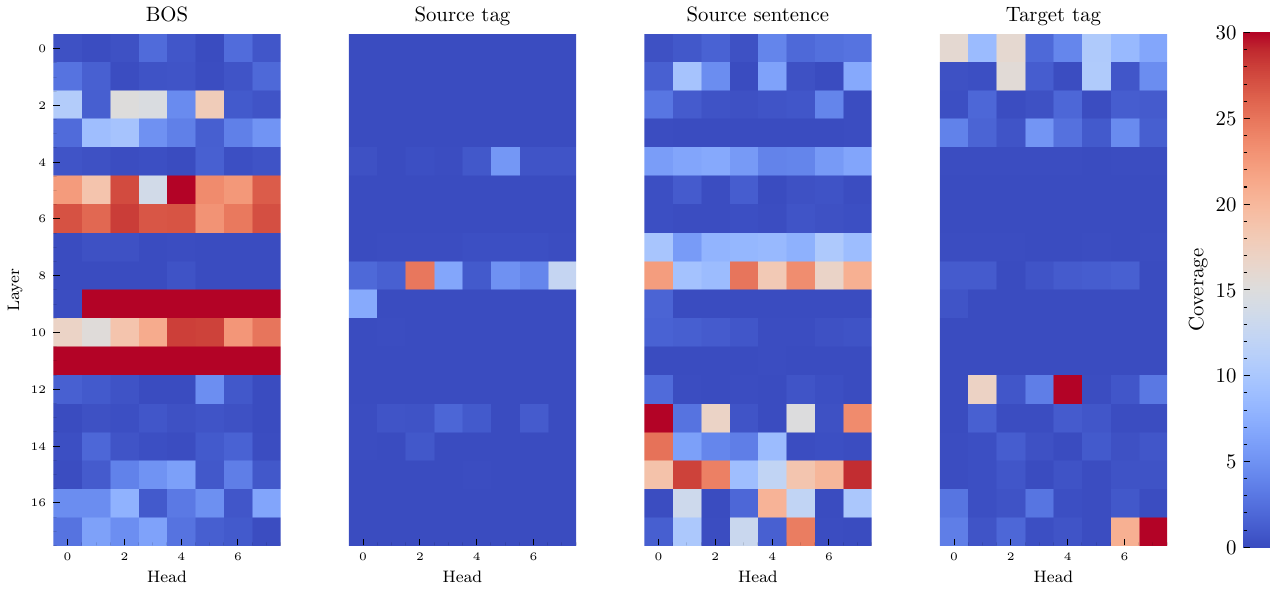}
  \caption {Coverage evaluating on \textsc{Flores-200} devtest using \parlam\ 128k. Each heatmap for each part of the prompt shows the coverage scores for each layer (vertical axis) and for each head (horizontal axis) in the model.  }\label{heatmapcoverage128}
\end{figure}

\begin{figure}[H]
  \centering
  \includegraphics[width=0.84\linewidth]{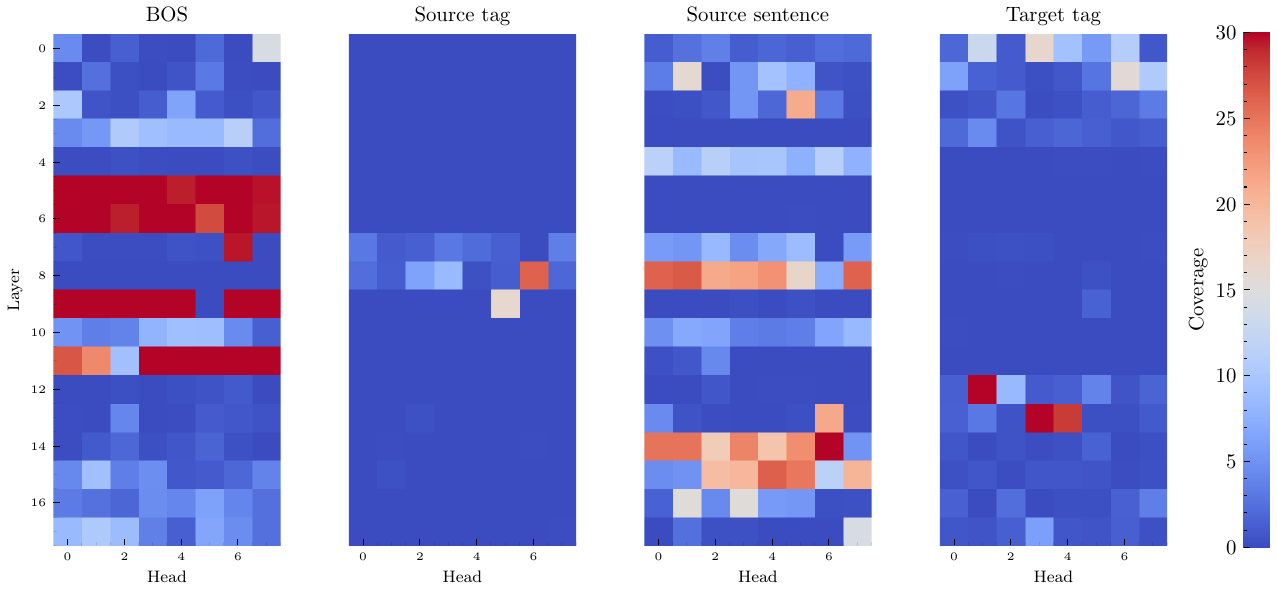}
  \caption {Coverage evaluating on \textsc{Flores-200} devtest using \parlam\ 256k. Each heatmap for each part of the prompt shows the coverage scores for each layer (vertical axis) and for each head (horizontal axis) in the model.  }\label{heatmapcoverage256}
\end{figure}

\begin{figure}[H]
  \includegraphics[width=\linewidth]{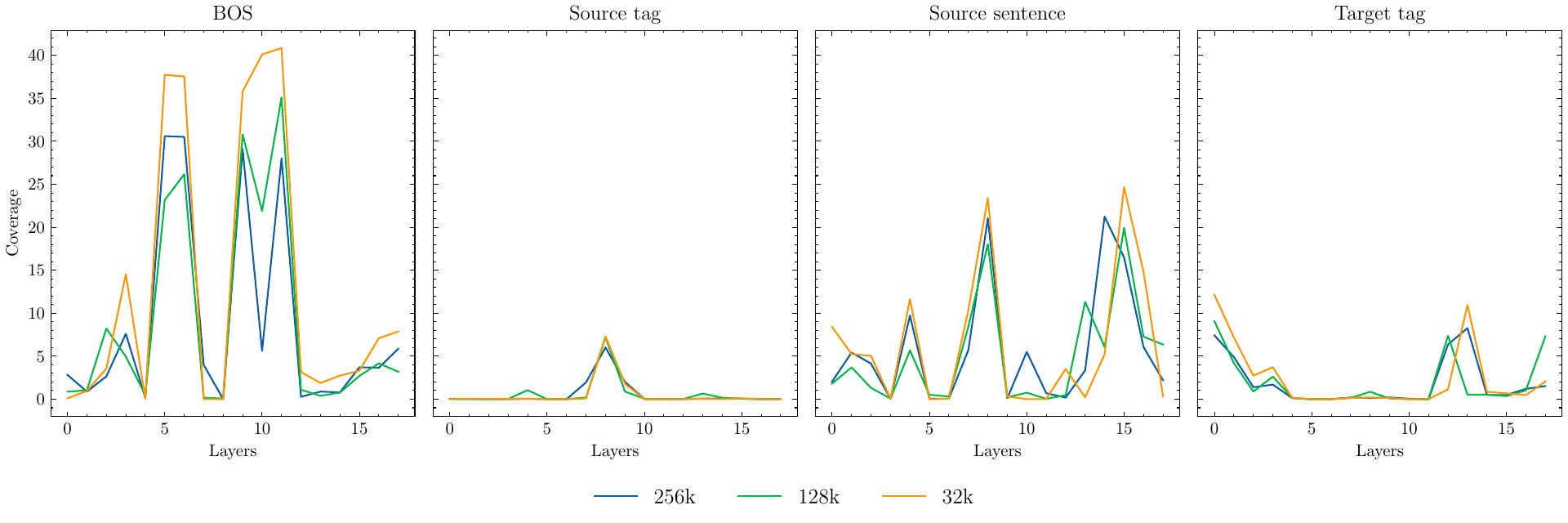}
  \caption {Average coverage per layer for each part of the prompt across various vocabulary sizes.}\label{comparison_coverage}
\end{figure}

We show in Figure \ref{heatmapcoverage128} and Figure \ref{heatmapcoverage256} the coverage heatmaps for \parlam\ 32k, 128k and 256k respectively. In Figure \ref{comparison_coverage} we show the average coverage per layer for the different vocabulary sizes. We notice that \parlam\ 32k, 128k and 256k exhibit a similar coverage pattern across layers.


\begin{figure}[H]
  \centering
  \includegraphics[width=0.84\linewidth]{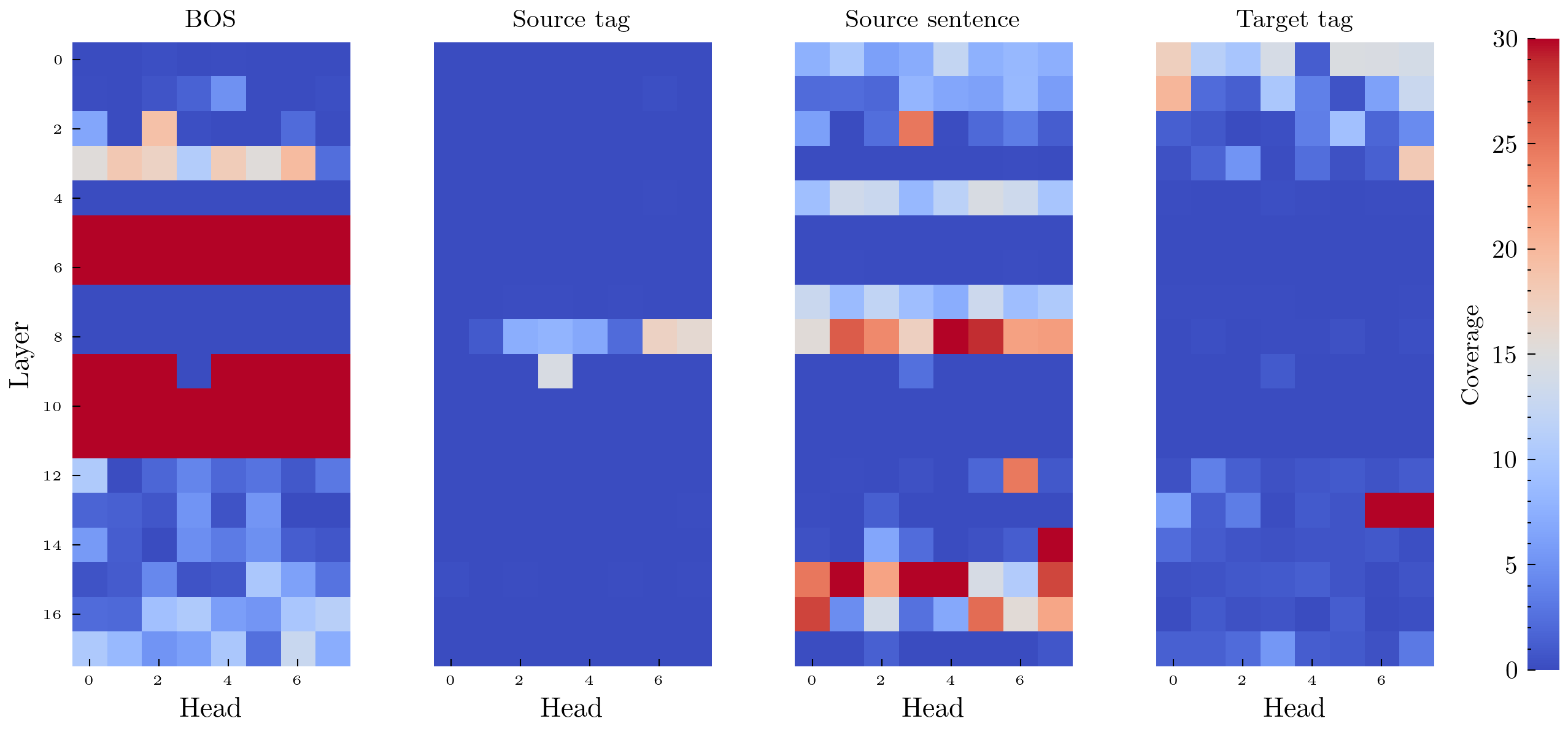}
  \caption {Coverage evaluating on \textsc{Flores-200} devtest using \parlam\ 32k. Each heatmap for each part of the prompt shows the coverage scores for each layer (vertical axis) and for each head (horizontal axis) in the model.  }\label{heatmapcoverage32_repeated}
\end{figure}

\subsection{Attention matrices}

\begin{figure}[H]
  \includegraphics[width=0.48\linewidth]{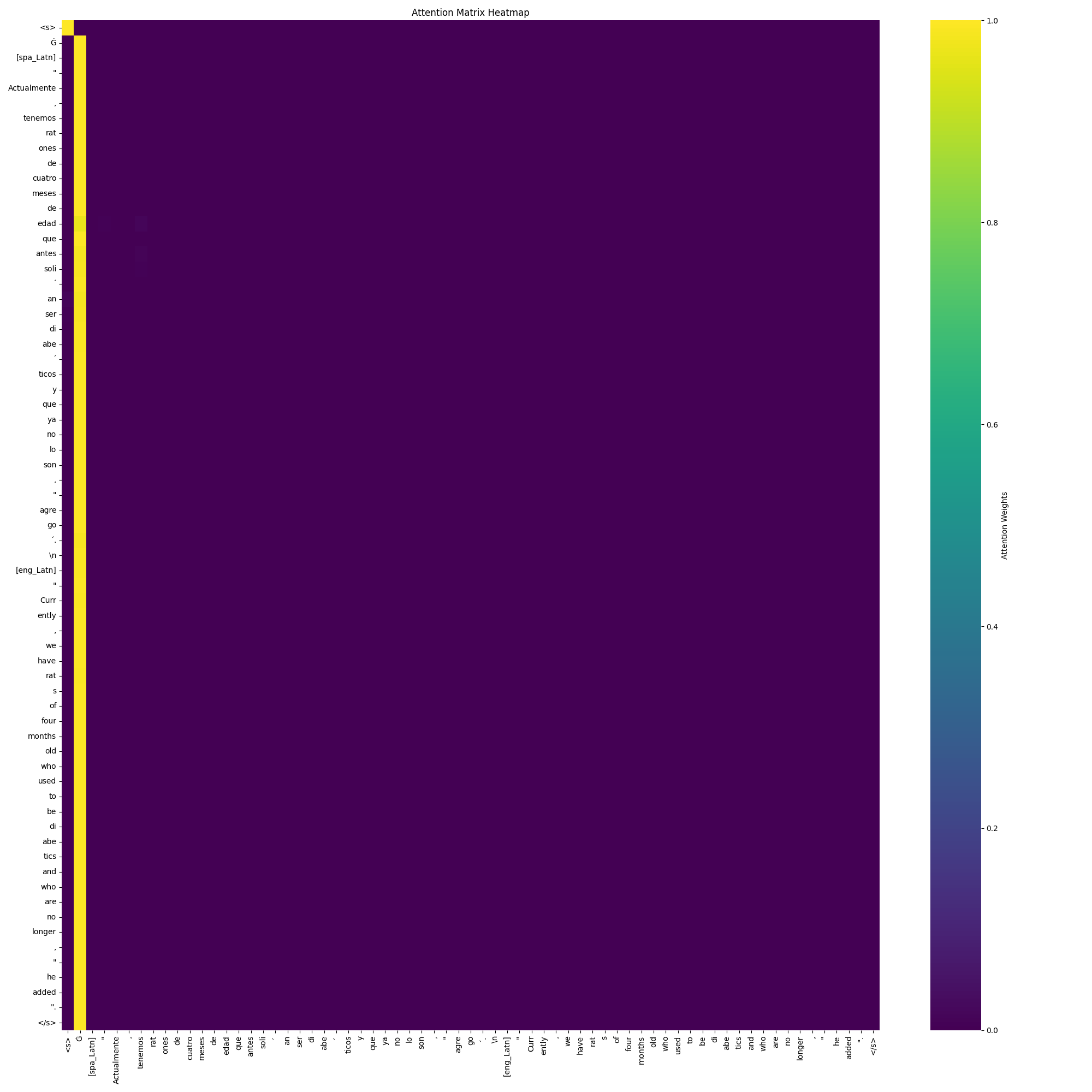} \hfill
  \includegraphics[width=0.48\linewidth]{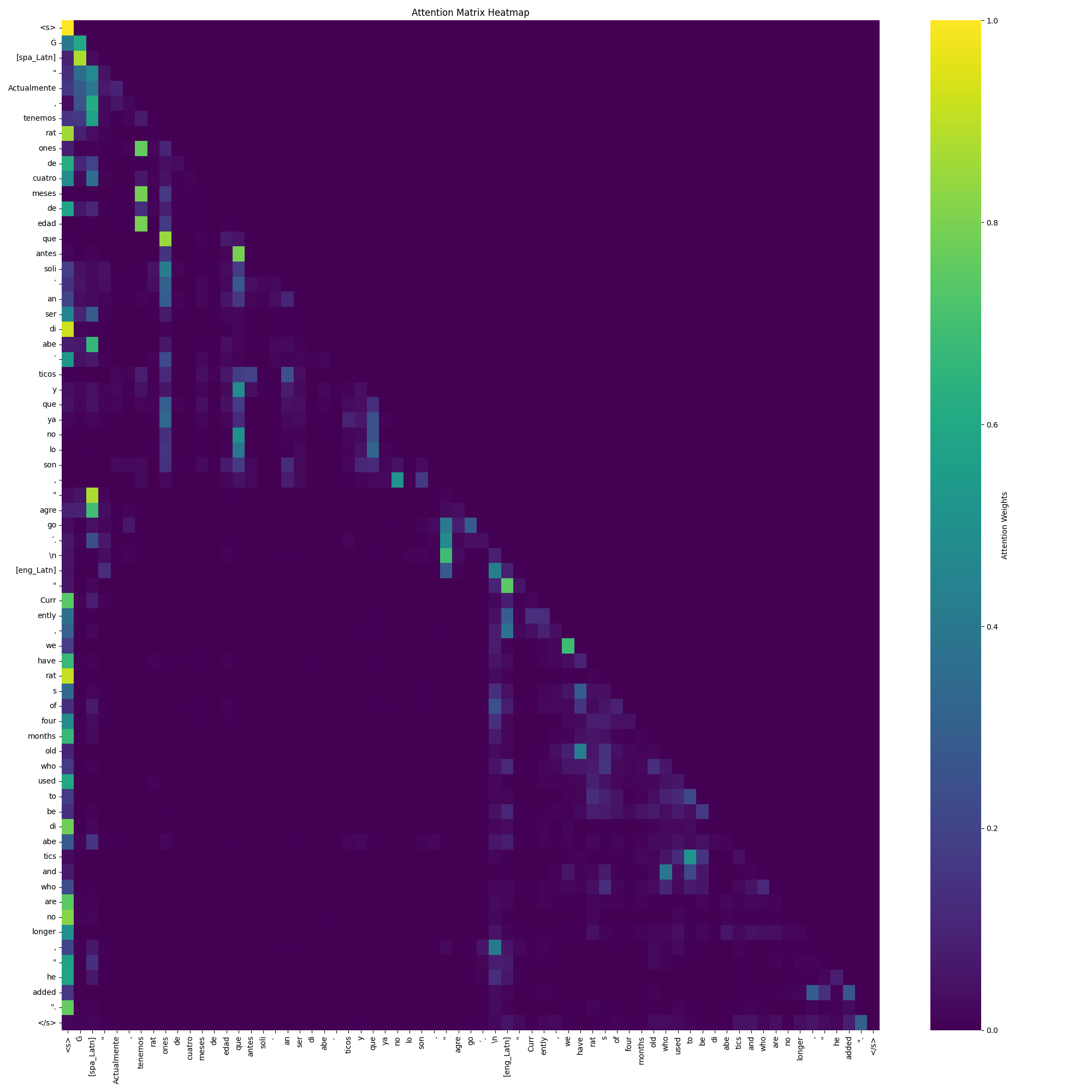}
  \caption {Attention weights for head 1 in layer 9 (left) and head 1 in layer 17 (right).}\label{fig:att_matrices}
\end{figure}

An attention sink mechanism occurs when all the attention mass is given to some special tokens. We visualize the attention matrices for the first head of layer 9 and layer 17 (last layer) in Figure \ref{fig:att_matrices}. We observe that in layer 9, the model is giving all the attention mass to the BOS token\footnote{ There is a special token created by Huggingface BPE implementation, which is positioned between the BOS and the source tag tokens. We consider this special token as part of the BOS token.} which allows the model to keep the residual stream of the network unchanged.

\section{Redundant Heads}\label{sec:redundant_heads}

Table \ref{tab:pruneheads} shows the number of heads that can be masked in various directions for different vocabulary sizes  without losing more than 2 BLEU points. We observe that as vocabulary size increases, we can mask more heads. Specifically, on average we can mask 41.56\%, 49.41\% and 64.7\% of the model's heads for \parlam\ 32k, \parlam\ 128k and \parlam\ 256k respectively.

\begin{table}[H]
  \centering
  \begin{tabular}{rrrr}
    \toprule
    & {\scriptsize \parlam \ \small 32k}  & {\scriptsize \parlam \ \small 128k} & {\scriptsize \parlam \ \small 256k} \\
    \midrule
    de$\rightarrow$ca & 64 & 64 & 88 \\
    de$\rightarrow$en & 32 & 72 & 88 \\
    de$\rightarrow$pt & 64 & 64 & 88 \\
    es$\rightarrow$ca & 64 & 104 & 88 \\
    es$\rightarrow$en & 64 & 72 & 88 \\
    es$\rightarrow$pt & 64 & 104 & 88 \\
    fr$\rightarrow$ca & 64 & 64 & 88 \\
    fr$\rightarrow$en & 24 & 72 & 88 \\
    fr$\rightarrow$pt & 64 & 0 & 88 \\
    gl$\rightarrow$ca & 64 & 104 & 88 \\
    gl$\rightarrow$en & 24 & 72 & 88 \\
    gl$\rightarrow$pt & 64 & 64 & 88 \\
    it$\rightarrow$ca & 64 & 80 & 88 \\
    it$\rightarrow$en & 64 & 72 & 88 \\
    it$\rightarrow$pt & 64 & 0 & 88 \\
    \cdashlinelr{1-4}
    \textbf{Avg.} & 56.53 & 67.2 & 88 \\
    \bottomrule
  \end{tabular}
  \caption{Number of masked heads across different language pairs and vocabulary sizes such that BLEU drop is less than 2 points.}
  \label{tab:pruneheads}
\end{table}

\section{Subspace distances}
\label{sec:subspace_distances}

\begin{figure}[H]
  \includegraphics[width=\linewidth]{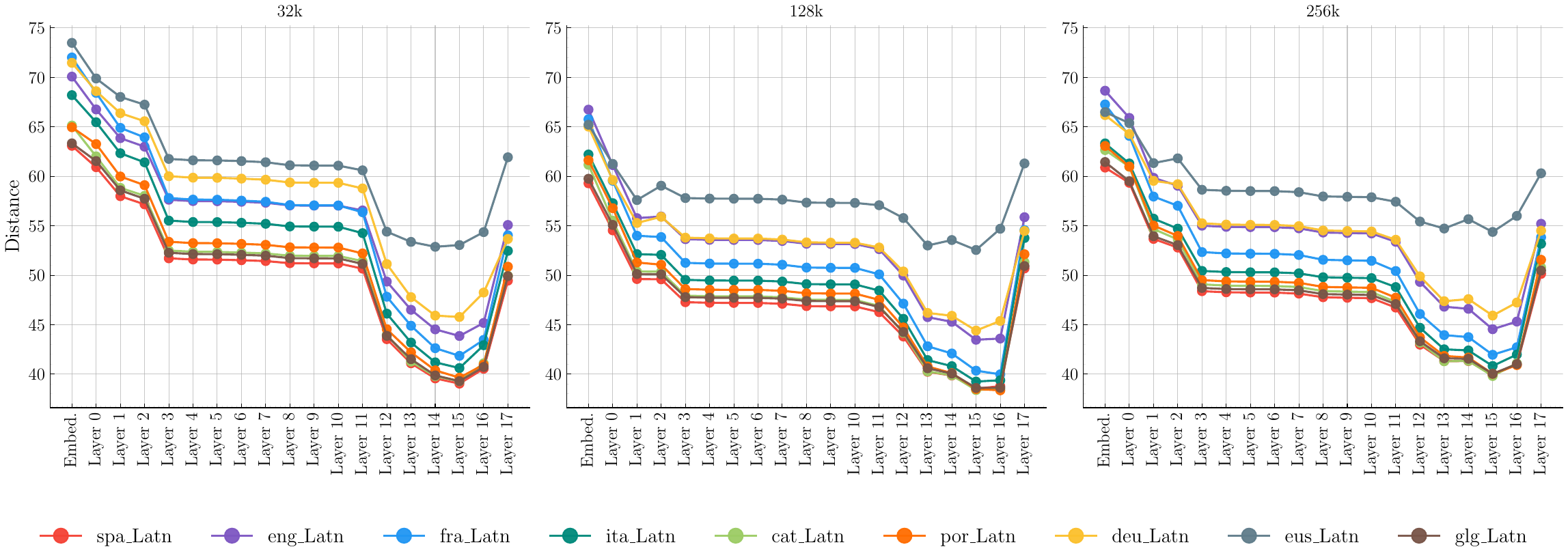}
  \caption{Mean distance between language subspaces grouped by languages and vocabulary sizes.}
  \label{fig:distances_mean_group_by_lang}
\end{figure}

We show in Figure \ref{fig:distances_mean_group_by_lang} the distances between language subspaces computed using the Riemannian metric on the space of positive definite matrices as detailed in \cite{chang-etal-2022-geometry} grouped by language and for each vocabulary size. We observe that for all the vocabulary sizes, Basque's subspace is further from the rest of the languages subspaces which could explain why model's performance on Basque is lower compared to other languages.

\section{UMAP Plots}
\label{sec:umap_plots_appendix}

Below we show the token representations\footnote{We use the first sentence from \flores\ devtest in each source language to construct the prompts: \texttt{"We now have 4-month-old mice that are non-diabetic that used to be diabetic," he added.}} using Uniform  Manifold Approximation and Projection (UMAP) \cite{mcinnes2018umap-software} for all the layers in \parlam\ 32k, 128k and 256k. We employ the cosine distance and we set the number of neighbours to 8 for computing UMAP's embeddings.

\begin{figure}[H]
  \includegraphics[width=\linewidth]{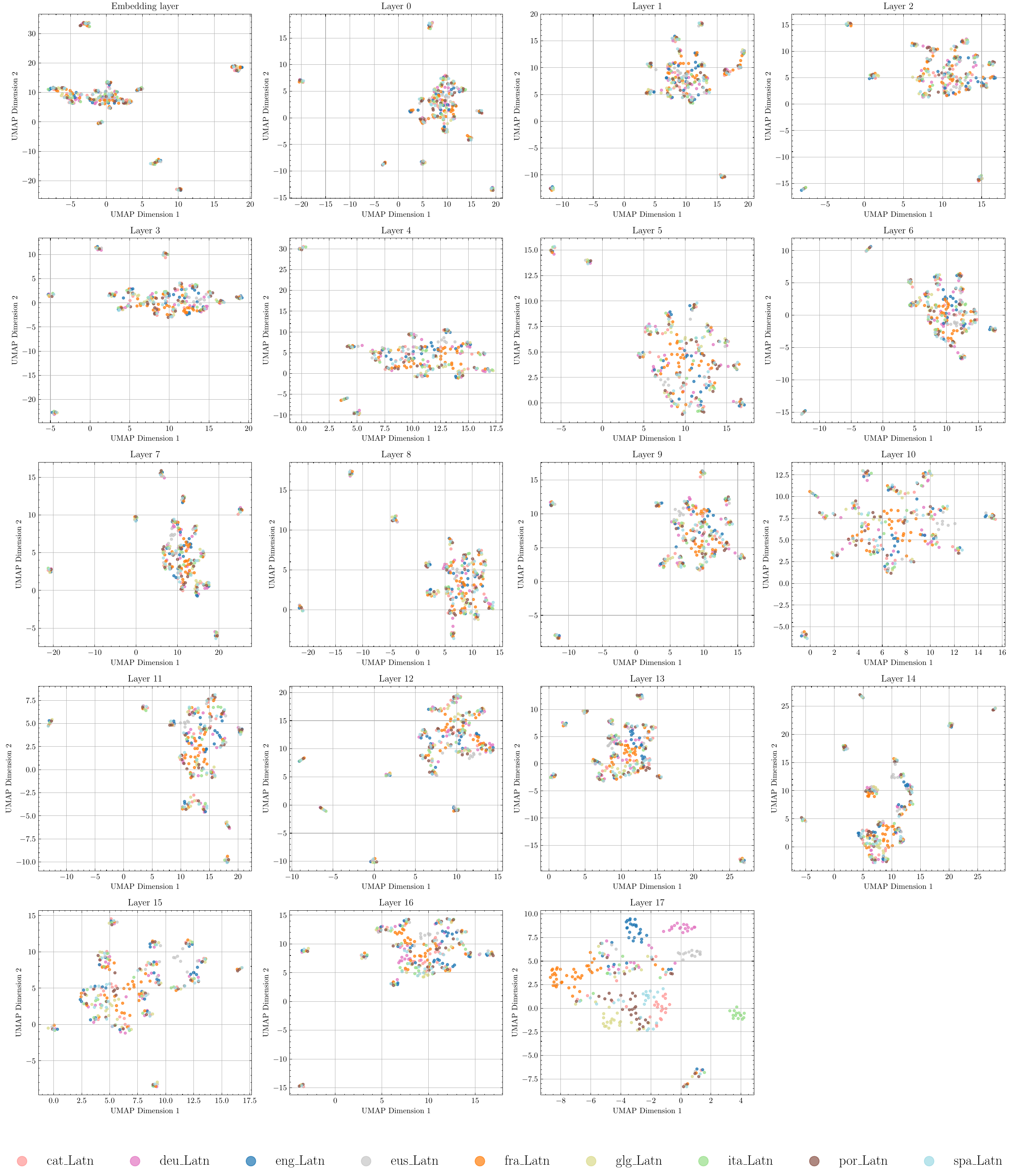}
  \caption{UMAP representations at the token embeddings in each layer grouped by source language using \parlam\ 32k.}
  \label{fig:umap_all_plots32}
\end{figure}

\begin{figure}[H]
  \includegraphics[width=\linewidth]{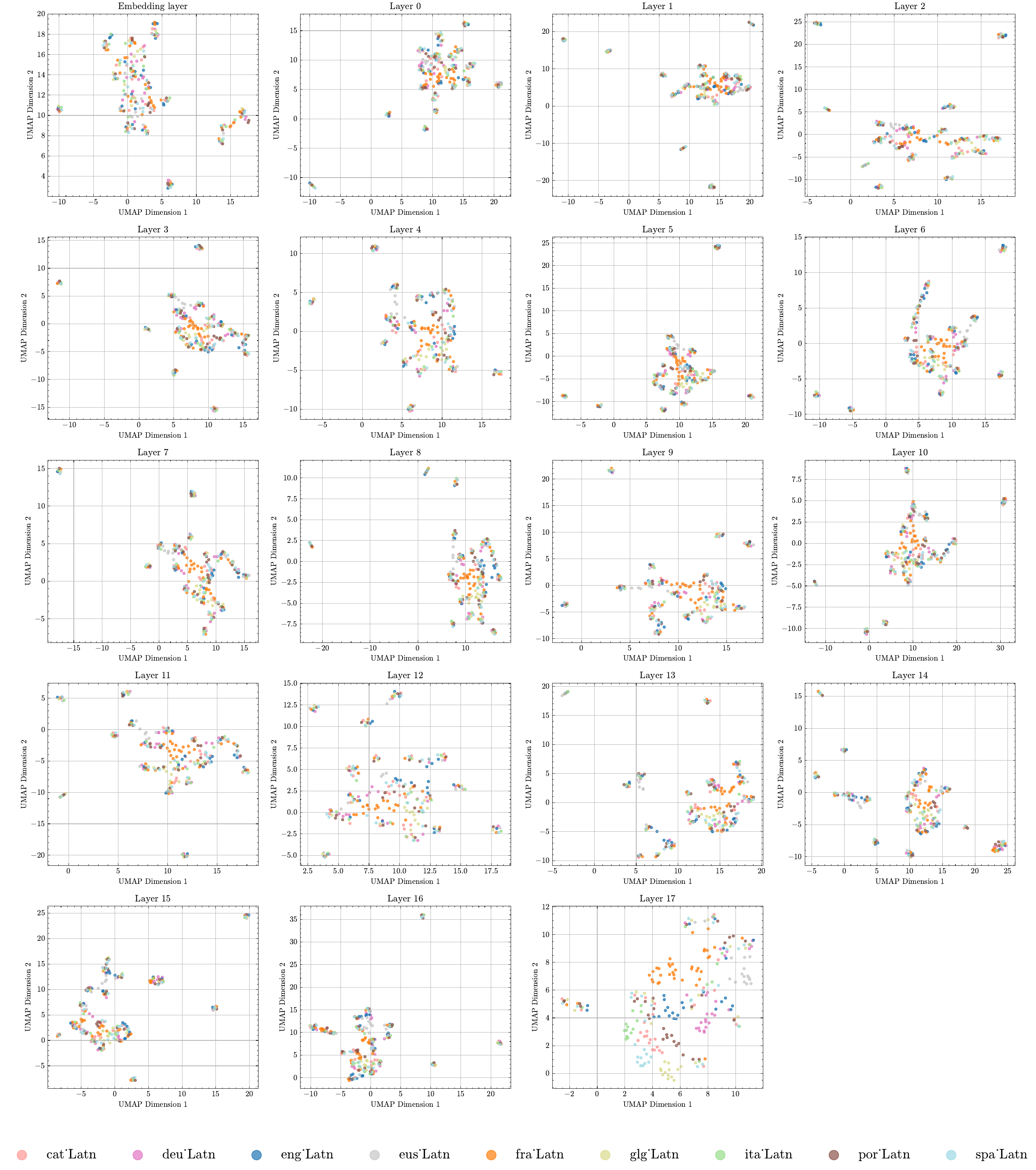}
  \caption{UMAP representations at the token embeddings in each layer grouped by source language using \parlam\ 128k.}
  \label{fig:umap_all_plots128}
\end{figure}

\begin{figure}[H]
  \includegraphics[width=\linewidth]{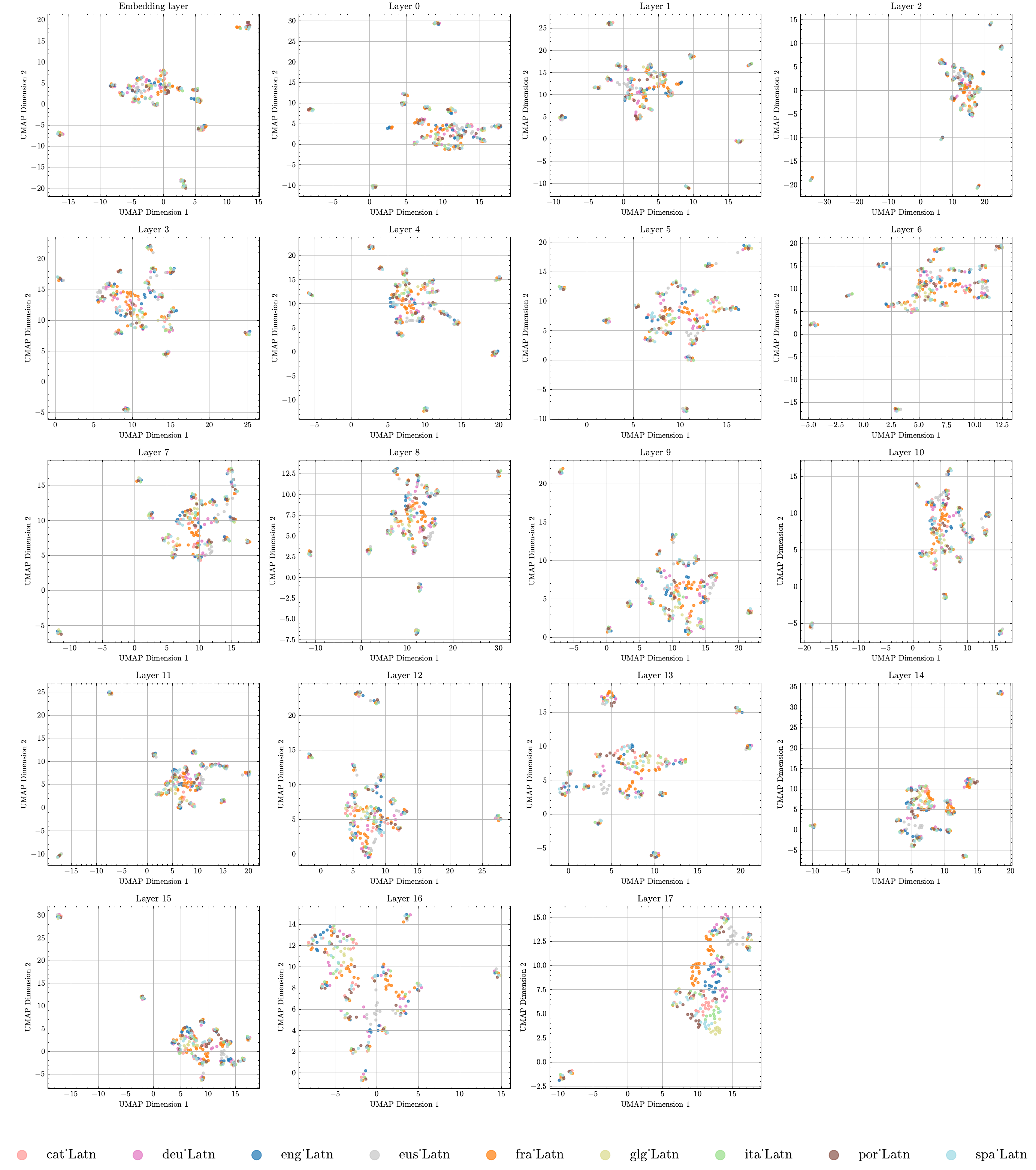}
  \caption{UMAP representations at the token embeddings in each layer grouped by source language using \parlam\ 256k.}
  \label{fig:umap_all_plots256}
\end{figure}

\subsection{Spherical Voronoi diagrams}\label{spherical_voronoi_app}

To better visualize high-dimensional token embeddings in \parlam\ models, we used spherical voronoi diagrams. Specifically, we reduced the embeddings to a 2D space, optimizing for cosine similarity using UMAP. Then, the 2D UMAP embeddings were projected onto a unit sphere. Specifically, each 2D point \((x, y)\) was mapped to 3D coordinates \((X, Y, Z)\)  as follows:

\begin{equation}
X = \sin(x) \cdot \cos(y) \;\;\;\;
Y = \sin(x) \cdot \sin(y) \;\;\;\;
Z = \cos(x)
\end{equation}

Then, for each language, we calculated the centroid of its corresponding tokens on the sphere and using these centroids, we computed Voronoi regions (where each region contains all the closest points to a specific centroid). We add as supplementary material the spherical voronoi diagrams for each layer in \parlam\ 32k.

\section{Detailed results}

We report in the following tables the results of \parlam\ models for each translation direction. We also provide comparisons for \textsc{TowerBase 7B} \cite{alves2024tower} in those directions that \parlam\ and \textsc{TowerBase 7B} share, as well as comparisons with \textsc{NLLB 3.3B} \cite{nllbteam2022language}.

\label{sec:detailed_results}

\begin{table*}[ht]
    \centering
    \small
    \caption{Results for ca$\rightarrow$xx.}
    \label{tab:results_caxx}
    \begin{tabular}{llrrrrrrrr} 
        \toprule
        & & \multicolumn{4}{c}{\flores} & \multicolumn{4}{c}{\ntrex} \\
        \cmidrule(lr){3-6} 	\cmidrule(lr){7-10}
        Pair & Model & {\footnotesize BLEU} & \textsc{ChrF} & COMET & \textsc{COMET-}\textsc{\tiny Kiwi} & BLEU & \textsc{ChrF} & COMET & \textsc{COMET-}\textsc{\tiny Kiwi} \\
        \midrule
        ca-de & \textsc{BSC} Bilinguals & 33.30 & 61.12 & 0.85 & 0.84 & 25.04 & 55.00 & 0.83 & 0.83 \\
         & \textsc{NLLB 3.3B}  & 31.19 & 58.41 & 0.85 & 0.84 & 21.72 & 53.41 & 0.81 & 0.82 \\
         \cdashlinelr{2-10}
         & \parlam\ 128k & 28.00 & 57.53 & 0.83 & 0.82 & 21.98 & 53.36 & 0.80 & 0.81 \\
         & \parlam\ 256k & 28.55 & 57.63 & 0.83 & 0.82 & 21.39 & 52.72 & 0.80 & 0.81 \\
         & \parlam\ 32k & 27.81 & 57.00 & 0.83 & 0.82 & 27.79 & 56.66 & 0.83 & 0.84 \\
        \cdashlinelr{1-10}
        ca-en & \textsc{BSC} Bilinguals & 46.29 & 70.44 & 0.88 & 0.86 & 41.20 & 66.57 & 0.87 & 0.86 \\
         & \textsc{NLLB 3.3B}  & 49.65 & 71.68 & 0.89 & 0.86 & 33.22 & 62.82 & 0.85 & 0.85 \\
         \cdashlinelr{2-10}
         & \parlam\ 128k & 42.91 & 68.69 & 0.88 & 0.86 & 33.73 & 63.07 & 0.85 & 0.85 \\
         & \parlam\ 256k & 42.47 & 68.47 & 0.88 & 0.85 & 32.82 & 62.14 & 0.85 & 0.84 \\
         & \parlam\ 32k & 41.92 & 68.15 & 0.87 & 0.85 & 37.61 & 64.98 & 0.87 & 0.85 \\
        \cdashlinelr{1-10}
        ca-es & \textsc{BSC} Bilinguals & 24.70 & 53.42 & 0.86 & 0.86 & 36.89 & 61.83 & 0.86 & 0.85 \\
         & \textsc{NLLB 3.3B}  & 25.62 & 53.73 & 0.86 & 0.86 & 35.44 & 61.27 & 0.86 & 0.85 \\
         \cdashlinelr{2-10}
         & \parlam\ 128k & 24.66 & 53.44 & 0.86 & 0.86 & 35.66 & 61.23 & 0.86 & 0.85 \\
         & \parlam\ 256k & 24.59 & 53.37 & 0.86 & 0.85 & 35.70 & 61.24 & 0.86 & 0.85 \\
         & \parlam\ 32k & 24.50 & 53.37 & 0.86 & 0.86 & 35.97 & 61.40 & 0.86 & 0.85 \\
        \cdashlinelr{1-10}
        ca-eu & \textsc{BSC} Bilinguals & 18.26 & 57.03 & 0.86 & 0.81 & 9.83 & 46.47 & 0.80 & 0.74 \\
         & \textsc{NLLB 3.3B}  & 13.13 & 50.47 & 0.83 & 0.75 & 12.40 & 49.99 & 0.82 & 0.78 \\
         \cdashlinelr{2-10}
         & \parlam\ 128k & 14.88 & 53.41 & 0.84 & 0.79 & 12.09 & 49.96 & 0.82 & 0.78 \\
         & \parlam\ 256k & 14.97 & 53.75 & 0.84 & 0.78 & 12.17 & 49.58 & 0.81 & 0.77 \\
         & \parlam\ 32k & 14.38 & 53.29 & 0.84 & 0.78 & 14.08 & 52.70 & 0.84 & 0.81 \\
        \cdashlinelr{1-10}
        ca-fr & \textsc{BSC} Bilinguals & 38.25 & 63.23 & 0.85 & 0.84 & 27.60 & 56.73 & 0.84 & 0.85 \\
         & \textsc{NLLB 3.3B}  & 39.89 & 64.05 & 0.86 & 0.85 & 25.20 & 54.13 & 0.81 & 0.82 \\
         \cdashlinelr{2-10}
         & \parlam\ 128k & 35.46 & 61.08 & 0.84 & 0.83 & 25.48 & 54.16 & 0.81 & 0.82 \\
         & \parlam\ 256k & 35.72 & 61.18 & 0.84 & 0.83 & 24.94 & 53.76 & 0.81 & 0.82 \\
         & \parlam\ 32k & 34.32 & 60.68 & 0.83 & 0.82 & 27.71 & 55.53 & 0.82 & 0.83 \\
        \cdashlinelr{1-10}
        ca-gl & \textsc{BSC} Bilinguals & 31.96 & 59.66 & 0.87 & 0.84 & 34.07 & 60.52 & 0.86 & 0.84 \\
         & \textsc{NLLB 3.3B}  & 32.78 & 59.25 & 0.87 & 0.85 & 33.23 & 60.22 & 0.86 & 0.84 \\
         \cdashlinelr{2-10}
         & \parlam\ 128k & 32.22 & 59.73 & 0.87 & 0.84 & 33.37 & 60.24 & 0.86 & 0.83 \\
         & \parlam\ 256k & 32.07 & 59.51 & 0.87 & 0.84 & 33.23 & 60.27 & 0.86 & 0.84 \\
         & \parlam\ 32k & 32.21 & 59.73 & 0.87 & 0.85 & 32.59 & 59.76 & 0.85 & 0.82 \\
        \cdashlinelr{1-10}
        ca-it & \textsc{BSC} Bilinguals & 26.92 & 56.55 & 0.87 & 0.85 & 29.46 & 58.00 & 0.87 & 0.85 \\
        & \textsc{NLLB 3.3B}  & 26.38 & 55.66 & 0.88 & 0.86 & 27.91 & 57.43 & 0.86 & 0.84 \\
        \cdashlinelr{2-10}
        & \parlam\ 128k & 25.77 & 55.78 & 0.87 & 0.85 & 28.11 & 57.62 & 0.86 & 0.84 \\
         & \parlam\ 256k & 25.76 & 55.94 & 0.87 & 0.85 & 27.80 & 57.33 & 0.85 & 0.84 \\
        & \parlam\ 32k & 25.45 & 55.51 & 0.87 & 0.85 & 29.07 & 57.95 & 0.86 & 0.84 \\
        \cdashlinelr{1-10}
        ca-pt & \textsc{BSC} Bilinguals & 37.18 & 62.73 & 0.88 & 0.84 & 31.46 & 57.67 & 0.86 & 0.84 \\
         & \textsc{NLLB 3.3B}  & 36.68 & 61.97 & 0.88 & 0.85 & 27.79 & 55.97 & 0.85 & 0.83 \\
         \cdashlinelr{2-10}
         & \parlam\ 128k & 36.27 & 62.12 & 0.88 & 0.84 & 28.50 & 56.29 & 0.85 & 0.83 \\
         & \parlam\ 256k & 35.76 & 61.88 & 0.88 & 0.84 & 27.92 & 55.91 & 0.85 & 0.83 \\
         & \parlam\ 32k & 35.81 & 61.67 & 0.88 & 0.84 & 28.19 & 56.17 & 0.85 & 0.83 \\
        \bottomrule
    \end{tabular}
\end{table*}

\begin{table*}[ht]
    \centering
    \small
    \caption{Results for de$\rightarrow$xx.}
    \label{tab:results_dexx}
    \begin{tabular}{llrrrrrrrr} 
        \toprule
        & & \multicolumn{4}{c}{\flores} & \multicolumn{4}{c}{\ntrex} \\
        \cmidrule(lr){3-6} 	\cmidrule(lr){7-10}
        Pair & Model & {\footnotesize BLEU} & \textsc{ChrF} & COMET & \textsc{COMET-}\textsc{\tiny Kiwi} & BLEU & \textsc{ChrF} & COMET & \textsc{COMET-}\textsc{\tiny Kiwi} \\
            \midrule
    de-ca & \textsc{BSC} Bilinguals & 30.15 & 57.65 & 0.83 & 0.82 & 28.24 & 55.02 & 0.83 & 0.84 \\
     & \textsc{NLLB 3.3B}  & 31.45 & 57.99 & 0.86 & 0.85 & 28.34 & 55.03 & 0.82 & 0.82 \\
     \cdashlinelr{2-10}
      & \parlam\ 128k & 32.23 & 59.02 & 0.85 & 0.83 & 28.13 & 54.66 & 0.82 & 0.82 \\
      & \parlam\ 256k & 31.76 & 58.73 & 0.85 & 0.83 & 27.94 & 54.58 & 0.82 & 0.81 \\
      & \parlam\ 32k & 31.76 & 58.56 & 0.85 & 0.83 & 24.49 & 53.60 & 0.78 & 0.80 \\
    \cdashlinelr{1-10}
    de-en & \textsc{NLLB 3.3B}  & 46.02 & 69.30 & 0.90 & 0.85 & 41.01 & 66.16 & 0.88 & 0.84 \\
      & \textsc{TowerBase 7B} & 43.69 & 68.56 & 0.89 & 0.84 & 41.01 & 66.16 & 0.88 & 0.84 \\
      \cdashlinelr{2-10}
      & \parlam\ 128k & 36.17 & 63.49 & 0.86 & 0.82 & 29.73 & 59.26 & 0.84 & 0.81 \\
      & \parlam\ 256k & 36.99 & 64.04 & 0.87 & 0.83 & 29.80 & 59.39 & 0.84 & 0.81 \\
      & \parlam\ 32k & 34.12 & 62.13 & 0.86 & 0.81 & 28.73 & 58.11 & 0.83 & 0.80 \\
    \cdashlinelr{1-10}
    de-es  & \textsc{NLLB 3.3B}  & 23.86 & 51.39 & 0.84 & 0.86 & 31.13 & 57.36 & 0.84 & 0.85 \\
      & \textsc{TowerBase 7B} & 21.66 & 50.94 & 0.83 & 0.85 & 31.13 & 57.36 & 0.84 & 0.85 \\
      \cdashlinelr{2-10}
      & \parlam\ 128k & 22.00 & 50.41 & 0.82 & 0.83 & 28.41 & 54.92 & 0.81 & 0.82 \\
      & \parlam\ 256k & 22.35 & 50.80 & 0.82 & 0.83 & 28.76 & 54.89 & 0.81 & 0.82 \\
      & \parlam\ 32k & 20.90 & 49.74 & 0.82 & 0.82 & 27.83 & 54.18 & 0.81 & 0.81 \\
    \cdashlinelr{1-10}
    de-eu & \textsc{NLLB 3.3B}  & 9.83 & 45.23 & 0.78 & 0.71 & 7.83 & 41.70 & 0.76 & 0.69 \\
    \cdashlinelr{2-10}
      & \parlam\ 128k & 9.91 & 46.23 & 0.78 & 0.73 & 8.18 & 42.65 & 0.75 & 0.72 \\
      & \parlam\ 256k & 11.48 & 47.52 & 0.79 & 0.74 & 8.93 & 43.59 & 0.76 & 0.73 \\
      & \parlam\ 32k & 10.77 & 46.22 & 0.77 & 0.72 & 8.46 & 42.39 & 0.74 & 0.71 \\
    \cdashlinelr{1-10}
    de-fr & \textsc{NLLB 3.3B}  & 37.62 & 62.60 & 0.86 & 0.85 & 28.06 & 56.03 & 0.83 & 0.85 \\
      & \textsc{TowerBase 7B} & 34.84 & 61.23 & 0.85 & 0.85 & 28.06 & 56.03 & 0.83 & 0.85 \\
      \cdashlinelr{2-10}
      & \parlam\ 128k & 28.50 & 56.32 & 0.80 & 0.80 & 20.26 & 49.16 & 0.77 & 0.78 \\
      & \parlam\ 256k & 29.01 & 56.15 & 0.80 & 0.79 & 20.84 & 49.13 & 0.77 & 0.78 \\
      & \parlam\ 32k & 27.13 & 54.89 & 0.79 & 0.78 & 20.37 & 48.30 & 0.75 & 0.76 \\
    \cdashlinelr{1-10}
    de-gl & \textsc{NLLB 3.3B}  & 28.87 & 55.70 & 0.85 & 0.85 & 29.17 & 56.21 & 0.84 & 0.84 \\
    \cdashlinelr{2-10}
      & \parlam\ 128k & 26.01 & 54.15 & 0.83 & 0.83 & 24.55 & 52.87 & 0.81 & 0.81 \\
      & \parlam\ 256k & 25.20 & 53.46 & 0.83 & 0.82 & 24.87 & 52.86 & 0.81 & 0.81 \\
      & \parlam\ 32k & 25.31 & 53.11 & 0.82 & 0.82 & 24.11 & 51.92 & 0.80 & 0.80 \\
    \cdashlinelr{1-10}
    de-it & \textsc{NLLB 3.3B}  & 25.88 & 54.95 & 0.87 & 0.86 & 27.84 & 56.12 & 0.86 & 0.85 \\
      & \textsc{TowerBase 7B} & 24.73 & 54.26 & 0.86 & 0.85 & 27.84 & 56.12 & 0.86 & 0.85 \\
      \cdashlinelr{2-10}
      & \parlam\ 128k & 22.47 & 52.44 & 0.84 & 0.83 & 22.77 & 52.04 & 0.82 & 0.82 \\
      & \parlam\ 256k & 22.74 & 52.34 & 0.85 & 0.83 & 23.12 & 52.16 & 0.82 & 0.82 \\
      & \parlam\ 32k & 21.36 & 51.19 & 0.84 & 0.82 & 22.39 & 51.53 & 0.81 & 0.81 \\
    \cdashlinelr{1-10}
    de-pt & \textsc{NLLB 3.3B}  & 33.42 & 59.32 & 0.87 & 0.85 & 29.42 & 55.97 & 0.85 & 0.85 \\
      & \textsc{TowerBase 7B} & 30.94 & 58.48 & 0.86 & 0.85 & 29.42 & 55.97 & 0.85 & 0.85 \\
      \cdashlinelr{2-10}
      & \parlam\ 128k & 30.02 & 57.17 & 0.85 & 0.83 & 24.09 & 51.90 & 0.82 & 0.82 \\
      & \parlam\ 256k & 30.36 & 57.46 & 0.85 & 0.83 & 24.06 & 51.90 & 0.82 & 0.82 \\
      & \parlam\ 32k & 29.19 & 55.98 & 0.84 & 0.81 & 23.00 & 51.09 & 0.80 & 0.80 \\
        \bottomrule
    \end{tabular}
\end{table*}

\begin{table*}[ht]
    \centering
    \small
    \caption{Results for en$\rightarrow$xx.}
    \label{tab:results_enxx}
    \begin{tabular}{llrrrrrrrr} 
        \toprule
        & & \multicolumn{4}{c}{\flores} & \multicolumn{4}{c}{\ntrex} \\
        \cmidrule(lr){3-6} 	\cmidrule(lr){7-10}
        Pair & Model & {\footnotesize BLEU} & \textsc{ChrF} & COMET & \textsc{COMET-}\textsc{\tiny Kiwi} & BLEU & \textsc{ChrF} & COMET & \textsc{COMET-}\textsc{\tiny Kiwi} \\
            \midrule
        en-ca & \textsc{BSC} Bilinguals & 44.05 & 67.95 & 0.88 & 0.85 & 37.49 & 62.38 & 0.85 & 0.83 \\
         & \textsc{NLLB 3.3B}  & 42.33 & 65.97 & 0.88 & 0.85 & 35.80 & 61.29 & 0.83 & 0.81 \\
         \cdashlinelr{2-10}
         & \parlam\ 128k & 42.29 & 66.44 & 0.87 & 0.84 & 35.95 & 61.30 & 0.83 & 0.81 \\
         & \parlam\ 256k & 42.64 & 66.59 & 0.87 & 0.84 & 35.05 & 60.72 & 0.82 & 0.80 \\
         & \parlam\ 32k & 42.32 & 66.39 & 0.86 & 0.84 & 37.93 & 63.19 & 0.84 & 0.82 \\
        \cdashlinelr{1-10}
        en-de & \textsc{NLLB 3.3B}  & 39.88 & 65.14 & 0.88 & 0.84 & 32.46 & 60.93 & 0.85 & 0.84 \\
         & \textsc{TowerBase 7B} & 37.53 & 64.47 & 0.87 & 0.84 & 32.46 & 60.93 & 0.85 & 0.84 \\
         \cdashlinelr{2-10}
         & \parlam\ 128k & 31.27 & 59.30 & 0.82 & 0.80 & 24.31 & 54.33 & 0.78 & 0.77 \\
         & \parlam\ 256k & 31.81 & 60.17 & 0.83 & 0.81 & 24.94 & 55.13 & 0.79 & 0.78 \\
         & \parlam\ 32k & 29.86 & 58.22 & 0.82 & 0.79 & 23.46 & 53.42 & 0.77 & 0.75 \\
        \cdashlinelr{1-10}
        en-es & \textsc{NLLB 3.3B}  & 28.14 & 55.85 & 0.86 & 0.86 & 39.33 & 63.79 & 0.85 & 0.84 \\
         & \textsc{TowerBase 7B} & 26.38 & 55.02 & 0.86 & 0.86 & 39.33 & 63.79 & 0.85 & 0.84 \\
         \cdashlinelr{2-10}
         & \parlam\ 128k & 24.34 & 53.01 & 0.83 & 0.84 & 35.62 & 60.75 & 0.81 & 0.80 \\
         & \parlam\ 256k & 25.00 & 53.43 & 0.84 & 0.84 & 36.42 & 61.36 & 0.82 & 0.81 \\
         & \parlam\ 32k & 23.47 & 52.61 & 0.83 & 0.83 & 34.86 & 60.10 & 0.81 & 0.79 \\
        \cdashlinelr{1-10}
        en-eu & \textsc{NLLB 3.3B}  & 15.71 & 53.25 & 0.85 & 0.82 & 11.62 & 47.74 & 0.81 & 0.79 \\
        \cdashlinelr{2-10}
         & \parlam\ 128k & 13.02 & 48.69 & 0.81 & 0.78 & 10.51 & 44.21 & 0.76 & 0.75 \\
         & \parlam\ 256k & 12.95 & 50.05 & 0.81 & 0.79 & 10.96 & 45.41 & 0.77 & 0.75 \\
         & \parlam\ 32k & 13.03 & 48.89 & 0.80 & 0.78 & 10.73 & 44.79 & 0.75 & 0.74 \\
        \cdashlinelr{1-10}
        en-fr & \textsc{NLLB 3.3B}  & 50.90 & 71.70 & 0.88 & 0.87 & 34.77 & 61.69 & 0.84 & 0.85 \\
         & \textsc{TowerBase 7B} & 49.28 & 70.83 & 0.88 & 0.87 & 34.77 & 61.69 & 0.84 & 0.85 \\
         \cdashlinelr{2-10}
         & \parlam\ 128k & 36.49 & 62.25 & 0.82 & 0.82 & 26.36 & 54.27 & 0.77 & 0.79 \\
         & \parlam\ 256k & 38.27 & 63.03 & 0.83 & 0.83 & 27.20 & 54.95 & 0.77 & 0.79 \\
         & \parlam\ 32k & 36.11 & 61.92 & 0.81 & 0.81 & 26.36 & 54.15 & 0.76 & 0.78 \\
        \cdashlinelr{1-10}
        en-gl & \textsc{NLLB 3.3B}  & 35.98 & 61.55 & 0.87 & 0.85 & 39.01 & 63.75 & 0.85 & 0.83 \\
        \cdashlinelr{2-10}
         & \parlam\ 128k & 32.26 & 59.64 & 0.85 & 0.83 & 33.28 & 59.53 & 0.81 & 0.79 \\
         & \parlam\ 256k & 32.61 & 59.66 & 0.85 & 0.83 & 33.13 & 59.59 & 0.81 & 0.79 \\
         & \parlam\ 32k & 31.16 & 58.92 & 0.84 & 0.82 & 31.88 & 58.48 & 0.80 & 0.77 \\
        \cdashlinelr{1-10}
        en-it & \textsc{NLLB 3.3B}  & 30.63 & 59.52 & 0.88 & 0.87 & 37.68 & 63.84 & 0.87 & 0.85 \\
        & \textsc{TowerBase 7B} & 29.64 & 59.13 & 0.88 & 0.87 & 37.68 & 63.84 & 0.87 & 0.85 \\
        \cdashlinelr{2-10}
         & \parlam\ 128k & 25.58 & 55.15 & 0.84 & 0.84 & 28.84 & 57.37 & 0.82 & 0.81 \\
         & \parlam\ 256k & 25.64 & 55.75 & 0.85 & 0.85 & 30.73 & 58.42 & 0.82 & 0.81 \\
         & \parlam\ 32k & 24.51 & 54.69 & 0.84 & 0.84 & 29.55 & 57.32 & 0.81 & 0.80 \\
        \cdashlinelr{1-10}
        en-pt & \textsc{NLLB 3.3B}  & 49.45 & 70.54 & 0.90 & 0.85 & 37.37 & 62.46 & 0.87 & 0.84 \\
         & \textsc{TowerBase 7B} & 49.67 & 71.36 & 0.90 & 0.85 & 37.37 & 62.46 & 0.87 & 0.84 \\
         \cdashlinelr{2-10}
         & \parlam\ 128k & 40.94 & 65.75 & 0.87 & 0.83 & 30.59 & 57.41 & 0.82 & 0.79 \\
         & \parlam\ 256k & 42.62 & 66.47 & 0.87 & 0.83 & 31.27 & 57.81 & 0.82 & 0.79 \\
         & \parlam\ 32k & 40.57 & 65.13 & 0.86 & 0.82 & 30.13 & 56.87 & 0.81 & 0.78 \\

        \bottomrule
    \end{tabular}
\end{table*}

\begin{table*}[ht]
    \centering
    \small
    \caption{Results for es$\rightarrow$xx.}
    \label{tab:results_esxx}
    \begin{tabular}{llrrrrrrrr} 
        \toprule
        & & \multicolumn{4}{c}{\flores} & \multicolumn{4}{c}{\ntrex} \\
        \cmidrule(lr){3-6} 	\cmidrule(lr){7-10}
        Pair & Model & {\footnotesize BLEU} & \textsc{ChrF} & COMET & \textsc{COMET-}\textsc{\tiny Kiwi} & BLEU & \textsc{ChrF} & COMET & \textsc{COMET-}\textsc{\tiny Kiwi} \\
            \midrule
    es-ca & \textsc{BSC} Bilinguals & 23.34 & 53.98 & 0.86 & 0.84 & 34.47 & 60.52 & 0.86 & 0.84 \\
     & \textsc{NLLB 3.3B}  & 25.70 & 55.24 & 0.86 & 0.84 & 33.16 & 60.59 & 0.86 & 0.83 \\
     \cdashlinelr{2-10}
     & \parlam\ 128k & 23.43 & 54.22 & 0.86 & 0.84 & 33.41 & 60.49 & 0.86 & 0.83 \\
     & \parlam\ 256k & 23.42 & 54.20 & 0.86 & 0.84 & 33.23 & 60.60 & 0.86 & 0.83 \\
     & \parlam\ 32k & 23.55 & 54.30 & 0.86 & 0.84 & 34.14 & 60.73 & 0.86 & 0.83 \\
     \cdashlinelr{1-10}
    es-de & \textsc{NLLB 3.3B}  & 22.88 & 53.27 & 0.84 & 0.84 & 24.63 & 55.15 & 0.83 & 0.84 \\
    & \textsc{TowerBase 7B}  & 18.86 & 51.44 & 0.82 & 0.84 & 24.63 & 55.15 & 0.83 & 0.84 \\
    \cdashlinelr{2-10}
    & \parlam\ 128k & 17.69 & 50.73 & 0.80 & 0.81 & 19.90 & 52.08 & 0.79 & 0.81 \\
     & \parlam\ 256k & 18.06 & 51.26 & 0.81 & 0.82 & 20.41 & 52.30 & 0.80 & 0.81 \\
     & \parlam\ 32k & 17.63 & 50.19 & 0.80 & 0.80 & 19.47 & 51.49 & 0.78 & 0.80 \\
     \cdashlinelr{1-10}
    es-en & \textsc{NLLB 3.3B}  & 32.93 & 61.52 & 0.88 & 0.86 & 41.88 & 67.47 & 0.88 & 0.86 \\
     & \textsc{TowerBase 7B}  & 30.47 & 60.37 & 0.87 & 0.86 & 41.88 & 67.47 & 0.88 & 0.86 \\
     \cdashlinelr{2-10}
     & \parlam\ 128k & 24.74 & 56.76 & 0.85 & 0.85 & 31.64 & 62.07 & 0.85 & 0.84 \\
     & \parlam\ 256k & 24.91 & 57.16 & 0.85 & 0.85 & 31.53 & 62.24 & 0.85 & 0.84 \\
     & \parlam\ 32k & 23.79 & 56.29 & 0.84 & 0.85 & 31.05 & 61.38 & 0.85 & 0.84 \\
     \cdashlinelr{1-10}
    es-eu & \textsc{NLLB 3.3B}  & 11.31 & 49.93 & 0.84 & 0.81 & 11.13 & 47.56 & 0.81 & 0.77 \\
    \cdashlinelr{2-10}
     & \parlam\ 128k & 10.39 & 49.12 & 0.82 & 0.81 & 11.45 & 48.54 & 0.81 & 0.79 \\
     & \parlam\ 256k & 11.22 & 49.59 & 0.83 & 0.81 & 11.29 & 48.92 & 0.81 & 0.79 \\
     & \parlam\ 32k & 11.26 & 49.16 & 0.82 & 0.79 & 11.31 & 47.79 & 0.80 & 0.78 \\
     \cdashlinelr{1-10}
    es-fr & \textsc{NLLB 3.3B}  & 29.97 & 58.18 & 0.85 & 0.86 & 27.92 & 56.77 & 0.84 & 0.85 \\
    & \textsc{TowerBase 7B}  & 25.16 & 55.84 & 0.84 & 0.85 & 27.92 & 56.77 & 0.84 & 0.85 \\
    \cdashlinelr{2-10}
     & \parlam\ 128k & 21.91 & 52.76 & 0.81 & 0.82 & 23.99 & 52.86 & 0.80 & 0.81 \\
     & \parlam\ 256k & 22.15 & 52.87 & 0.81 & 0.82 & 23.85 & 52.99 & 0.80 & 0.81 \\
     & \parlam\ 32k & 21.96 & 52.78 & 0.81 & 0.82 & 24.39 & 53.10 & 0.79 & 0.81 \\
     \cdashlinelr{1-10}
    es-gl & \textsc{NLLB 3.3B}  & 24.64 & 53.77 & 0.87 & 0.84 & 34.92 & 61.24 & 0.87 & 0.83 \\
    \cdashlinelr{2-10}
     & \parlam\ 128k & 21.47 & 52.69 & 0.87 & 0.84 & 33.34 & 60.71 & 0.86 & 0.83 \\
     & \parlam\ 256k & 21.59 & 52.54 & 0.86 & 0.84 & 33.63 & 60.81 & 0.86 & 0.82 \\
     & \parlam\ 32k & 21.29 & 52.51 & 0.86 & 0.84 & 33.08 & 60.63 & 0.86 & 0.83 \\
     \cdashlinelr{1-10}
    es-it & \textsc{NLLB 3.3B}  & 22.77 & 52.86 & 0.87 & 0.86 & 29.60 & 58.19 & 0.87 & 0.85 \\
    & \textsc{TowerBase 7B}  & 19.95 & 51.18 & 0.86 & 0.86 & 29.60 & 58.19 & 0.87 & 0.85 \\
    \cdashlinelr{2-10}
     & \parlam\ 128k & 18.76 & 50.27 & 0.85 & 0.85 & 25.08 & 55.31 & 0.84 & 0.83 \\
     & \parlam\ 256k & 18.86 & 50.53 & 0.85 & 0.84 & 25.42 & 55.57 & 0.85 & 0.84 \\
     & \parlam\ 32k & 19.29 & 50.45 & 0.85 & 0.84 & 25.14 & 55.55 & 0.84 & 0.83 \\
     \cdashlinelr{1-10}
    es-pt & \textsc{NLLB 3.3B}  & 26.18 & 55.23 & 0.87 & 0.85 & 32.30 & 58.24 & 0.87 & 0.84 \\
     & \textsc{TowerBase 7B}  & 23.11 & 53.87 & 0.87 & 0.85 & 32.30 & 58.24 & 0.87 & 0.84 \\
     \cdashlinelr{2-10}
     & \parlam\ 128k & 21.16 & 52.25 & 0.86 & 0.84 & 25.82 & 54.84 & 0.85 & 0.83 \\
     & \parlam\ 256k & 21.84 & 52.70 & 0.86 & 0.84 & 27.27 & 55.53 & 0.85 & 0.83 \\
     & \parlam\ 32k & 21.65 & 52.74 & 0.86 & 0.84 & 27.00 & 55.35 & 0.85 & 0.83 \\

        \bottomrule
    \end{tabular}
\end{table*}

\begin{table*}[ht]
    \centering
    \small
    \caption{Results for eu$\rightarrow$xx.}
    \label{tab:results_euxx}
    \begin{tabular}{llrrrrrrrr} 
        \toprule
        & & \multicolumn{4}{c}{\flores} & \multicolumn{4}{c}{\ntrex} \\
        \cmidrule(lr){3-6} 	\cmidrule(lr){7-10}
        Pair & Model & {\footnotesize BLEU} & \textsc{ChrF} & COMET & \textsc{COMET-}\textsc{\tiny Kiwi} & BLEU & \textsc{ChrF} & COMET & \textsc{COMET-}\textsc{\tiny Kiwi} \\
            \midrule
        eu-ca & \textsc{BSC} Bilinguals & 26.18 & 54.14 & 0.85 & 0.82 & 24.56 & 51.56 & 0.83 & 0.81 \\
         & \textsc{NLLB 3.3B}  & 26.70 & 53.97 & 0.86 & 0.82 & 22.29 & 49.79 & 0.81 & 0.79 \\
         \cdashlinelr{2-10}
         & \parlam\ 128k & 24.33 & 51.85 & 0.84 & 0.80 & 21.70 & 49.48 & 0.81 & 0.78 \\
         & \parlam\ 256k & 24.02 & 51.67 & 0.84 & 0.80 & 20.19 & 48.69 & 0.80 & 0.77 \\
         & \parlam\ 32k & 22.92 & 50.69 & 0.83 & 0.79 & 24.29 & 51.84 & 0.82 & 0.81 \\
        \cdashlinelr{1-10}
        eu-de & \textsc{NLLB 3.3B}  & 22.71 & 51.75 & 0.83 & 0.80 & 18.96 & 48.84 & 0.81 & 0.79 \\
        \cdashlinelr{2-10}
         & \parlam\ 128k & 13.64 & 44.72 & 0.76 & 0.72 & 11.38 & 41.74 & 0.73 & 0.72 \\
         & \parlam\ 256k & 13.58 & 44.77 & 0.76 & 0.72 & 10.74 & 41.78 & 0.73 & 0.72 \\
         & \parlam\ 32k & 10.62 & 40.74 & 0.72 & 0.69 & 9.30 & 38.93 & 0.69 & 0.69 \\
        \cdashlinelr{1-10}
        eu-en & \textsc{NLLB 3.3B}  & 33.44 & 60.57 & 0.87 & 0.86 & 29.59 & 57.37 & 0.85 & 0.85 \\
        \cdashlinelr{2-10}
         & \parlam\ 128k & 21.49 & 51.65 & 0.82 & 0.81 & 16.70 & 48.58 & 0.79 & 0.80 \\
         & \parlam\ 256k & 22.12 & 52.31 & 0.82 & 0.82 & 16.41 & 48.54 & 0.79 & 0.80 \\
         & \parlam\ 32k & 17.52 & 48.60 & 0.79 & 0.78 & 13.84 & 45.54 & 0.77 & 0.77 \\
        \cdashlinelr{1-10}
        eu-es & \textsc{NLLB 3.3B}  & 20.50 & 48.29 & 0.84 & 0.84 & 27.50 & 53.84 & 0.84 & 0.83 \\
        \cdashlinelr{2-10}
         & \parlam\ 128k & 17.74 & 45.98 & 0.81 & 0.81 & 20.71 & 48.75 & 0.79 & 0.79 \\
         & \parlam\ 256k & 17.94 & 45.41 & 0.81 & 0.81 & 20.58 & 48.54 & 0.79 & 0.79 \\
         & \parlam\ 32k & 15.61 & 43.47 & 0.79 & 0.79 & 18.76 & 47.03 & 0.78 & 0.78 \\
        \cdashlinelr{1-10}
        eu-fr & \textsc{NLLB 3.3B}  & 29.05 & 56.00 & 0.84 & 0.83 & 22.63 & 50.58 & 0.81 & 0.82 \\
        \cdashlinelr{2-10}
         & \parlam\ 128k & 18.58 & 46.77 & 0.75 & 0.75 & 14.90 & 42.94 & 0.73 & 0.73 \\
         & \parlam\ 256k & 18.39 & 46.08 & 0.75 & 0.74 & 14.73 & 42.58 & 0.72 & 0.72 \\
         & \parlam\ 32k & 15.77 & 44.00 & 0.71 & 0.71 & 12.58 & 40.59 & 0.69 & 0.70 \\
        \cdashlinelr{1-10}
        eu-gl & \textsc{NLLB 3.3B}  & 25.16 & 52.52 & 0.86 & 0.83 & 24.18 & 52.15 & 0.83 & 0.82 \\
        \cdashlinelr{2-10}
         & \parlam\ 128k & 19.24 & 47.58 & 0.82 & 0.78 & 18.04 & 46.91 & 0.79 & 0.77 \\
         & \parlam\ 256k & 18.53 & 46.92 & 0.81 & 0.78 & 18.23 & 46.74 & 0.79 & 0.76 \\
         & \parlam\ 32k & 15.91 & 45.11 & 0.79 & 0.75 & 16.13 & 44.99 & 0.77 & 0.75 \\
        \cdashlinelr{1-10}
        eu-it & \textsc{NLLB 3.3B}  & 21.27 & 51.07 & 0.86 & 0.84 & 22.45 & 51.13 & 0.84 & 0.83 \\
        \cdashlinelr{2-10}
         & \parlam\ 128k & 16.39 & 45.65 & 0.81 & 0.80 & 16.82 & 46.45 & 0.79 & 0.79 \\
         & \parlam\ 256k & 16.46 & 45.76 & 0.81 & 0.80 & 15.96 & 46.05 & 0.79 & 0.78 \\
         & \parlam\ 32k & 14.01 & 43.52 & 0.79 & 0.77 & 14.34 & 44.19 & 0.77 & 0.76 \\
        \cdashlinelr{1-10}
        eu-pt & \textsc{NLLB 3.3B}  & 27.79 & 54.65 & 0.86 & 0.84 & 23.93 & 50.72 & 0.83 & 0.82 \\
        \cdashlinelr{2-10}
         & \parlam\ 128k & 20.12 & 48.58 & 0.82 & 0.80 & 16.11 & 44.79 & 0.79 & 0.78 \\
         & \parlam\ 256k & 20.89 & 48.87 & 0.81 & 0.80 & 16.80 & 45.27 & 0.79 & 0.78 \\
         & \parlam\ 32k & 17.64 & 46.34 & 0.79 & 0.77 & 14.05 & 42.96 & 0.76 & 0.76 \\
        \bottomrule
    \end{tabular}
\end{table*}

\begin{table*}[ht]
    \centering
    \small
    \caption{Results for fr$\rightarrow$xx.}
    \label{tab:results_frxx}
    \begin{tabular}{llrrrrrrrr} 
        \toprule
        & & \multicolumn{4}{c}{\flores} & \multicolumn{4}{c}{\ntrex} \\
        \cmidrule(lr){3-6} 	\cmidrule(lr){7-10}
        Pair & Model & {\footnotesize BLEU} & \textsc{ChrF} & COMET & \textsc{COMET-}\textsc{\tiny Kiwi} & BLEU & \textsc{ChrF} & COMET & \textsc{COMET-}\textsc{\tiny Kiwi} \\
            \midrule
    
    fr-ca & \textsc{BSC} Bilinguals & 34.44 & 60.10 & 0.86 & 0.83 & 29.22 & 55.76 & 0.84 & 0.83 \\
     & \textsc{NLLB 3.3B}  & 34.00 & 59.82 & 0.87 & 0.84 & 27.30 & 54.40 & 0.83 & 0.82 \\
     \cdashlinelr{2-10}
     & \parlam\ 128k & 34.35 & 60.24 & 0.86 & 0.83 & 27.57 & 54.40 & 0.83 & 0.81 \\
     & \parlam\ 256k & 33.63 & 59.83 & 0.86 & 0.83 & 27.00 & 54.18 & 0.83 & 0.81 \\
     & \parlam\ 32k & 34.28 & 60.16 & 0.86 & 0.83 & 27.03 & 54.04 & 0.83 & 0.81 \\
     \cdashlinelr{1-10}
    fr-de & \textsc{NLLB 3.3B}  & 29.96 & 57.73 & 0.85 & 0.84 & 23.82 & 53.55 & 0.83 & 0.84 \\
     & \textsc{TowerBase 7B}  & 25.48 & 56.02 & 0.82 & 0.84 & 23.82 & 53.55 & 0.83 & 0.84 \\
     \cdashlinelr{2-10}
     & \parlam\ 128k & 24.63 & 54.96 & 0.81 & 0.80 & 19.07 & 49.59 & 0.78 & 0.78 \\
     & \parlam\ 256k & 23.85 & 54.54 & 0.82 & 0.80 & 18.18 & 49.18 & 0.78 & 0.78 \\
     & \parlam\ 32k & 22.45 & 53.56 & 0.81 & 0.78 & 18.35 & 48.80 & 0.77 & 0.77 \\
     \cdashlinelr{1-10}
    fr-en & \textsc{NLLB 3.3B}  & 48.38 & 70.72 & 0.90 & 0.86 & 40.30 & 64.78 & 0.87 & 0.86 \\
     & \textsc{TowerBase 7B}  & 45.48 & 69.54 & 0.89 & 0.86 & 40.30 & 64.78 & 0.87 & 0.86 \\
     \cdashlinelr{2-10}
     & \parlam\ 128k & 37.37 & 64.47 & 0.87 & 0.85 & 28.95 & 58.15 & 0.84 & 0.84 \\
     & \parlam\ 256k & 37.74 & 64.80 & 0.87 & 0.85 & 29.11 & 58.37 & 0.84 & 0.84 \\
     & \parlam\ 32k & 34.87 & 63.11 & 0.86 & 0.84 & 28.36 & 57.38 & 0.83 & 0.83 \\
     \cdashlinelr{1-10}
    fr-es & \textsc{NLLB 3.3B}  & 24.45 & 52.39 & 0.86 & 0.86 & 32.28 & 57.85 & 0.85 & 0.85 \\
     & \textsc{TowerBase 7B}  & 22.02 & 51.42 & 0.84 & 0.85 & 32.28 & 57.85 & 0.85 & 0.85 \\
     \cdashlinelr{2-10}
     & \parlam\ 128k & 21.65 & 50.63 & 0.84 & 0.84 & 27.18 & 54.18 & 0.82 & 0.83 \\
     & \parlam\ 256k & 21.80 & 50.74 & 0.84 & 0.84 & 27.30 & 54.22 & 0.82 & 0.83 \\
     & \parlam\ 32k & 21.60 & 50.66 & 0.84 & 0.84 & 27.23 & 54.00 & 0.82 & 0.82 \\
     \cdashlinelr{1-10}
    fr-eu & \textsc{NLLB 3.3B}  & 10.73 & 46.16 & 0.80 & 0.73 & 7.79 & 41.10 & 0.76 & 0.69 \\
    \cdashlinelr{2-10}
     & \parlam\ 128k & 10.79 & 48.17 & 0.80 & 0.76 & 9.32 & 44.51 & 0.78 & 0.75 \\
     & \parlam\ 256k & 11.78 & 48.71 & 0.80 & 0.77 & 9.43 & 44.37 & 0.78 & 0.75 \\
     & \parlam\ 32k & 11.59 & 48.08 & 0.79 & 0.75 & 8.65 & 43.30 & 0.76 & 0.72 \\
     \cdashlinelr{1-10}
    fr-gl & \textsc{NLLB 3.3B}  & 30.59 & 57.45 & 0.86 & 0.85 & 29.61 & 56.42 & 0.85 & 0.84 \\
    \cdashlinelr{2-10}
     & \parlam\ 128k & 27.95 & 55.92 & 0.85 & 0.84 & 24.65 & 52.84 & 0.81 & 0.81 \\
     & \parlam\ 256k & 28.49 & 55.94 & 0.85 & 0.84 & 24.57 & 52.94 & 0.82 & 0.81 \\
     & \parlam\ 32k & 27.69 & 55.65 & 0.85 & 0.83 & 24.11 & 52.42 & 0.81 & 0.81 \\
     \cdashlinelr{1-10}
    fr-it & \textsc{NLLB 3.3B}  & 27.06 & 56.27 & 0.88 & 0.86 & 28.22 & 56.47 & 0.86 & 0.86 \\
     & \textsc{TowerBase 7B}  & 25.14 & 55.00 & 0.87 & 0.86 & 28.22 & 56.47 & 0.86 & 0.86 \\
     \cdashlinelr{2-10}
     & \parlam\ 128k & 24.45 & 53.92 & 0.86 & 0.84 & 24.25 & 53.18 & 0.84 & 0.83 \\
     & \parlam\ 256k & 24.27 & 53.92 & 0.86 & 0.84 & 24.45 & 53.22 & 0.84 & 0.83 \\
     & \parlam\ 32k & 23.98 & 53.72 & 0.86 & 0.84 & 23.84 & 53.05 & 0.83 & 0.82 \\
     \cdashlinelr{1-10}
    fr-pt & \textsc{NLLB 3.3B}  & 36.18 & 61.28 & 0.88 & 0.85 & 29.11 & 55.64 & 0.85 & 0.84 \\
     & \textsc{TowerBase 7B}  & 33.03 & 60.10 & 0.87 & 0.85 & 29.11 & 55.64 & 0.85 & 0.84 \\
     \cdashlinelr{2-10}
     & \parlam\ 128k & 32.15 & 59.00 & 0.86 & 0.83 & 24.59 & 52.51 & 0.83 & 0.82 \\
     & \parlam\ 256k & 32.86 & 59.22 & 0.86 & 0.83 & 24.85 & 52.21 & 0.82 & 0.81 \\
     & \parlam\ 32k & 31.72 & 58.70 & 0.86 & 0.82 & 24.33 & 52.19 & 0.82 & 0.81 \\
        \bottomrule
    \end{tabular}
\end{table*}

\begin{table*}[ht]
    \centering
    \small
    \caption{Results for it$\rightarrow$xx.}
    \label{tab:results_itxx}
    \begin{tabular}{llrrrrrrrr} 
        \toprule
        & & \multicolumn{4}{c}{\flores} & \multicolumn{4}{c}{\ntrex} \\
        \cmidrule(lr){3-6} 	\cmidrule(lr){7-10}
        Pair & Model & {\footnotesize BLEU} & \textsc{ChrF} & COMET & \textsc{COMET-}\textsc{\tiny Kiwi} & BLEU & \textsc{ChrF} & COMET & \textsc{COMET-}\textsc{\tiny Kiwi} \\
            \midrule

        it-ca & \textsc{BSC} Bilinguals & 27.68 & 56.63 & 0.86 & 0.84 & 31.87 & 57.96 & 0.86 & 0.84 \\
         & \textsc{NLLB 3.3B}  & 27.77 & 56.56 & 0.87 & 0.86 & 31.18 & 57.64 & 0.85 & 0.83 \\
         \cdashlinelr{2-10}
         & \parlam\ 128k & 27.92 & 57.34 & 0.87 & 0.85 & 31.00 & 57.62 & 0.85 & 0.83 \\
         & \parlam\ 256k & 27.86 & 57.25 & 0.87 & 0.85 & 30.69 & 57.35 & 0.85 & 0.83 \\
         & \parlam\ 32k & 27.48 & 57.19 & 0.86 & 0.85 & 30.67 & 57.08 & 0.84 & 0.82 \\
        \cdashlinelr{1-10}
        it-de & \textsc{NLLB 3.3B}  & 25.33 & 55.23 & 0.85 & 0.86 & 26.76 & 56.82 & 0.84 & 0.85 \\
         & \textsc{TowerBase 7B} & 18.14 & 49.13 & 0.82 & 0.86 & 26.76 & 56.82 & 0.84 & 0.85 \\
         \cdashlinelr{2-10}
         & \parlam\ 128k & 20.84 & 52.75 & 0.82 & 0.83 & 20.84 & 51.69 & 0.79 & 0.82 \\
         & \parlam\ 256k & 21.05 & 53.04 & 0.82 & 0.83 & 21.06 & 52.07 & 0.80 & 0.82 \\
         & \parlam\ 32k & 19.77 & 51.78 & 0.81 & 0.82 & 20.28 & 51.35 & 0.79 & 0.80 \\
        \cdashlinelr{1-10}
        it-en & \textsc{NLLB 3.3B}  & 36.33 & 64.25 & 0.88 & 0.87 & 43.96 & 67.59 & 0.88 & 0.86 \\
         & \textsc{TowerBase 7B} & 32.95 & 62.57 & 0.88 & 0.86 & 43.96 & 67.59 & 0.88 & 0.86 \\
         \cdashlinelr{2-10}
         & \parlam\ 128k & 27.80 & 58.98 & 0.86 & 0.85 & 33.76 & 62.30 & 0.85 & 0.84 \\
         & \parlam\ 256k & 28.91 & 59.82 & 0.86 & 0.86 & 34.76 & 62.75 & 0.85 & 0.85 \\
         & \parlam\ 32k & 27.43 & 58.75 & 0.85 & 0.85 & 32.90 & 61.49 & 0.84 & 0.84 \\
        \cdashlinelr{1-10}
        it-es & \textsc{NLLB 3.3B}  & 22.70 & 51.45 & 0.86 & 0.87 & 34.15 & 59.45 & 0.86 & 0.86 \\
        & \textsc{TowerBase 7B} & 20.71 & 50.87 & 0.85 & 0.87 & 34.15 & 59.45 & 0.86 & 0.86 \\
        \cdashlinelr{2-10}
         & \parlam\ 128k & 20.91 & 50.70 & 0.85 & 0.86 & 30.30 & 56.88 & 0.84 & 0.85 \\
         & \parlam\ 256k & 21.35 & 51.04 & 0.85 & 0.86 & 30.62 & 56.96 & 0.84 & 0.85 \\
         & \parlam\ 32k & 20.99 & 50.72 & 0.85 & 0.86 & 30.06 & 56.70 & 0.84 & 0.85 \\
        \cdashlinelr{1-10}
        it-eu & \textsc{NLLB 3.3B}  & 7.65 & 43.50 & 0.79 & 0.73 & 8.09 & 41.63 & 0.76 & 0.70 \\
        \cdashlinelr{2-10}
         & \parlam\ 128k & 9.77 & 47.74 & 0.81 & 0.79 & 10.07 & 45.74 & 0.79 & 0.76 \\
         & \parlam\ 256k & 11.33 & 49.20 & 0.82 & 0.80 & 10.82 & 46.47 & 0.79 & 0.77 \\
         & \parlam\ 32k & 10.69 & 48.55 & 0.81 & 0.78 & 10.44 & 45.82 & 0.78 & 0.76 \\
        \cdashlinelr{1-10}
        it-fr & \textsc{NLLB 3.3B}  & 33.24 & 60.44 & 0.87 & 0.87 & 29.23 & 57.43 & 0.84 & 0.86 \\
         & \textsc{TowerBase 7B} & 29.16 & 58.49 & 0.85 & 0.87 & 29.23 & 57.43 & 0.84 & 0.86 \\
         \cdashlinelr{2-10}
         & \parlam\ 128k & 27.21 & 56.24 & 0.83 & 0.84 & 23.92 & 52.66 & 0.81 & 0.82 \\
         & \parlam\ 256k & 27.89 & 56.11 & 0.83 & 0.84 & 24.39 & 52.83 & 0.80 & 0.82 \\
         & \parlam\ 32k & 26.35 & 55.67 & 0.82 & 0.83 & 24.04 & 52.53 & 0.80 & 0.81 \\
        \cdashlinelr{1-10}
        it-gl & \textsc{NLLB 3.3B}  & 25.72 & 54.62 & 0.87 & 0.86 & 32.39 & 58.86 & 0.86 & 0.84 \\
        \cdashlinelr{2-10}
         & \parlam\ 128k & 23.80 & 54.06 & 0.86 & 0.85 & 29.04 & 56.66 & 0.84 & 0.83 \\
         & \parlam\ 256k & 23.79 & 53.94 & 0.86 & 0.84 & 29.34 & 56.60 & 0.84 & 0.82 \\
         & \parlam\ 32k & 23.59 & 53.88 & 0.85 & 0.84 & 28.20 & 55.97 & 0.84 & 0.82 \\
        \cdashlinelr{1-10}
        it-pt & \textsc{NLLB 3.3B}  & 28.17 & 56.94 & 0.88 & 0.86 & 33.41 & 58.86 & 0.87 & 0.85 \\
         & \textsc{TowerBase 7B} & 24.49 & 55.37 & 0.86 & 0.85 & 33.41 & 58.86 & 0.87 & 0.85 \\
         \cdashlinelr{2-10}
         & \parlam\ 128k & 26.64 & 56.24 & 0.87 & 0.84 & 28.48 & 55.43 & 0.85 & 0.83 \\
         & \parlam\ 256k & 27.10 & 56.52 & 0.87 & 0.85 & 28.33 & 55.31 & 0.84 & 0.83 \\
         & \parlam\ 32k & 25.86 & 55.58 & 0.86 & 0.84 & 28.03 & 55.24 & 0.84 & 0.82 \\
        \bottomrule
    \end{tabular}
\end{table*}

\begin{table*}[ht]
    \centering
    \small
    \caption{Results for gl$\rightarrow$xx.}
    \label{tab:results_glxx}
    \begin{tabular}{llrrrrrrrr} 
        \toprule
        & & \multicolumn{4}{c}{\flores} & \multicolumn{4}{c}{\ntrex} \\
        \cmidrule(lr){3-6} 	\cmidrule(lr){7-10}
        Pair & Model & {\footnotesize BLEU} & \textsc{ChrF} & COMET & \textsc{COMET-}\textsc{\tiny Kiwi} & BLEU & \textsc{ChrF} & COMET & \textsc{COMET-}\textsc{\tiny Kiwi} \\
            \midrule
        gl-ca & \textsc{BSC} Bilinguals & 32.43 & 60.50 & 0.87 & 0.84 & 34.27 & 60.27 & 0.86 & 0.84 \\
         & \textsc{NLLB 3.3B}  & 34.43 & 60.88 & 0.87 & 0.85 & 34.25 & 60.34 & 0.86 & 0.83 \\
         \cdashlinelr{2-10}
         & \parlam\ 128k & 32.77 & 60.71 & 0.87 & 0.84 & 34.28 & 60.55 & 0.86 & 0.83 \\
         & \parlam\ 256k & 33.00 & 60.85 & 0.88 & 0.84 & 34.10 & 60.42 & 0.86 & 0.83 \\
         & \parlam\ 32k & 32.75 & 60.76 & 0.87 & 0.84 & 33.72 & 60.27 & 0.86 & 0.83 \\
        \cdashlinelr{1-10}
        gl-de & \textsc{NLLB 3.3B}  & 29.57 & 57.53 & 0.85 & 0.84 & 25.13 & 55.12 & 0.83 & 0.83 \\
        \cdashlinelr{2-10}
         & \parlam\ 128k & 23.05 & 54.44 & 0.81 & 0.81 & 20.23 & 51.72 & 0.79 & 0.80 \\
         & \parlam\ 256k & 24.25 & 55.47 & 0.82 & 0.82 & 20.35 & 52.31 & 0.79 & 0.80 \\
         & \parlam\ 32k & 22.89 & 54.11 & 0.80 & 0.80 & 19.75 & 51.46 & 0.78 & 0.79 \\
        \cdashlinelr{1-10}
        gl-en & \textsc{NLLB 3.3B}  & 44.14 & 68.60 & 0.89 & 0.86 & 43.52 & 67.80 & 0.88 & 0.85 \\
        \cdashlinelr{2-10}
         & \parlam\ 128k & 35.47 & 64.50 & 0.86 & 0.85 & 33.40 & 62.42 & 0.85 & 0.84 \\
         & \parlam\ 256k & 34.74 & 64.17 & 0.86 & 0.84 & 32.56 & 62.21 & 0.85 & 0.84 \\
         & \parlam\ 32k & 34.15 & 63.48 & 0.86 & 0.84 & 30.76 & 61.22 & 0.84 & 0.83 \\
        \cdashlinelr{1-10}
        gl-es & \textsc{NLLB 3.3B}  & 25.59 & 53.47 & 0.87 & 0.85 & 36.99 & 61.92 & 0.87 & 0.84 \\
        \cdashlinelr{2-10}
         & \parlam\ 128k & 23.67 & 52.86 & 0.86 & 0.85 & 35.18 & 61.04 & 0.86 & 0.84 \\
         & \parlam\ 256k & 23.79 & 52.87 & 0.86 & 0.85 & 35.84 & 61.32 & 0.86 & 0.84 \\
         & \parlam\ 32k & 23.59 & 52.83 & 0.86 & 0.85 & 35.48 & 61.15 & 0.86 & 0.84 \\
        \cdashlinelr{1-10}
        gl-eu & \textsc{NLLB 3.3B}  & 12.37 & 48.45 & 0.82 & 0.73 & 9.06 & 43.94 & 0.78 & 0.70 \\
        \cdashlinelr{2-10}
         & \parlam\ 128k & 13.23 & 51.10 & 0.83 & 0.77 & 11.89 & 48.13 & 0.81 & 0.76 \\
         & \parlam\ 256k & 13.68 & 51.27 & 0.83 & 0.77 & 11.28 & 48.44 & 0.81 & 0.76 \\
         & \parlam\ 32k & 12.78 & 50.05 & 0.82 & 0.75 & 10.94 & 47.31 & 0.80 & 0.74 \\
        \cdashlinelr{1-10}
        gl-fr & \textsc{NLLB 3.3B}  & 38.37 & 63.38 & 0.86 & 0.85 & 29.03 & 56.98 & 0.84 & 0.84 \\
        \cdashlinelr{2-10}
         & \parlam\ 128k & 29.14 & 57.49 & 0.82 & 0.82 & 23.19 & 52.26 & 0.79 & 0.81 \\
         & \parlam\ 256k & 30.24 & 57.82 & 0.82 & 0.82 & 23.80 & 52.55 & 0.79 & 0.80 \\
         & \parlam\ 32k & 29.84 & 57.65 & 0.81 & 0.81 & 23.56 & 52.22 & 0.79 & 0.80 \\
        \cdashlinelr{1-10}
        gl-it & \textsc{NLLB 3.3B}  & 26.14 & 55.52 & 0.88 & 0.85 & 30.79 & 58.39 & 0.87 & 0.84 \\
        \cdashlinelr{2-10}
         & \parlam\ 128k & 22.73 & 53.29 & 0.86 & 0.84 & 26.47 & 55.68 & 0.84 & 0.83 \\
         & \parlam\ 256k & 23.20 & 53.77 & 0.86 & 0.84 & 27.00 & 56.19 & 0.84 & 0.83 \\
         & \parlam\ 32k & 22.45 & 53.22 & 0.86 & 0.84 & 26.36 & 55.84 & 0.84 & 0.83 \\
        \cdashlinelr{1-10}
        gl-pt & \textsc{NLLB 3.3B}  & 34.42 & 60.37 & 0.88 & 0.83 & 31.87 & 58.16 & 0.87 & 0.83 \\
        \cdashlinelr{2-10}
         & \parlam\ 128k & 28.42 & 57.24 & 0.87 & 0.83 & 26.36 & 54.81 & 0.85 & 0.81 \\
         & \parlam\ 256k & 29.11 & 57.70 & 0.87 & 0.83 & 27.82 & 55.65 & 0.85 & 0.81 \\
         & \parlam\ 32k & 29.23 & 57.83 & 0.87 & 0.83 & 27.50 & 55.41 & 0.85 & 0.81 \\

        \bottomrule
    \end{tabular}
\end{table*}

\begin{table*}[ht]
    \centering
    \small
    \caption{Results for pt$\rightarrow$xx.}
    \label{tab:results_ptxx}
    \begin{tabular}{llrrrrrrrr} 
        \toprule
        & & \multicolumn{4}{c}{\flores} & \multicolumn{4}{c}{\ntrex} \\
        \cmidrule(lr){3-6} 	\cmidrule(lr){7-10}
        Pair & Model & {\footnotesize BLEU} & \textsc{ChrF} & COMET & \textsc{COMET-}\textsc{\tiny Kiwi} & BLEU & \textsc{ChrF} & COMET & \textsc{COMET-}\textsc{\tiny Kiwi} \\
            \midrule

        pt-ca & \textsc{BSC} Bilinguals & 35.75 & 61.22 & 0.87 & 0.84 & 32.04 & 58.28 & 0.86 & 0.83 \\
         & \textsc{NLLB 3.3B}  & 34.64 & 60.68 & 0.87 & 0.84 & 31.17 & 57.91 & 0.85 & 0.83 \\
         \cdashlinelr{2-10}
         & \parlam\ 128k & 35.50 & 61.41 & 0.87 & 0.84 & 31.05 & 57.84 & 0.85 & 0.83 \\
         & \parlam\ 256k & 35.38 & 60.95 & 0.87 & 0.83 & 31.12 & 57.84 & 0.85 & 0.83 \\
         & \parlam\ 32k & 35.50 & 61.26 & 0.87 & 0.83 & 30.95 & 57.66 & 0.85 & 0.82 \\
        \cdashlinelr{1-10}
        pt-de & \textsc{NLLB 3.3B}  & 31.27 & 58.75 & 0.85 & 0.85 & 25.56 & 55.62 & 0.84 & 0.84 \\
         & \textsc{TowerBase 7B} & 25.48 & 56.02 & 0.82 & 0.84 & 25.56 & 55.62 & 0.84 & 0.84 \\
         \cdashlinelr{2-10}
         & \parlam\ 128k & 25.45 & 55.44 & 0.82 & 0.82 & 19.99 & 51.73 & 0.80 & 0.80 \\
         & \parlam\ 256k & 26.51 & 55.90 & 0.83 & 0.82 & 20.03 & 51.96 & 0.80 & 0.81 \\
         & \parlam\ 32k & 25.01 & 54.48 & 0.81 & 0.81 & 20.48 & 51.29 & 0.79 & 0.79 \\
        \cdashlinelr{1-10}
        pt-en & \textsc{NLLB 3.3B}  & 52.50 & 73.31 & 0.90 & 0.85 & 43.94 & 68.11 & 0.88 & 0.85 \\
         & \textsc{TowerBase 7B} & 50.16 & 72.76 & 0.90 & 0.85 & 43.94 & 68.11 & 0.88 & 0.85 \\
         \cdashlinelr{2-10}
         & \parlam\ 128k & 42.71 & 68.42 & 0.88 & 0.84 & 33.21 & 62.26 & 0.85 & 0.83 \\
         & \parlam\ 256k & 43.31 & 68.95 & 0.88 & 0.84 & 33.50 & 62.46 & 0.86 & 0.83 \\
         & \parlam\ 32k & 41.73 & 67.58 & 0.87 & 0.83 & 32.87 & 61.63 & 0.85 & 0.82 \\
        \cdashlinelr{1-10}
        pt-es & \textsc{NLLB 3.3B}  & 25.76 & 53.31 & 0.86 & 0.86 & 34.85 & 60.45 & 0.86 & 0.85 \\
         & \textsc{TowerBase 7B} & 22.82 & 51.90 & 0.85 & 0.85 & 34.85 & 60.45 & 0.86 & 0.85 \\
         \cdashlinelr{2-10}
         & \parlam\ 128k & 22.97 & 51.85 & 0.85 & 0.85 & 30.89 & 57.40 & 0.85 & 0.84 \\
         & \parlam\ 256k & 23.04 & 51.82 & 0.85 & 0.84 & 31.32 & 57.66 & 0.85 & 0.84 \\
         & \parlam\ 32k & 22.72 & 51.74 & 0.85 & 0.84 & 30.84 & 57.25 & 0.85 & 0.84 \\
        \cdashlinelr{1-10}
        pt-eu & \textsc{NLLB 3.3B}  & 10.38 & 45.45 & 0.79 & 0.72 & 8.14 & 41.30 & 0.76 & 0.69 \\
        \cdashlinelr{2-10}
         & \parlam\ 128k & 11.18 & 49.09 & 0.82 & 0.79 & 9.93 & 46.18 & 0.80 & 0.77 \\
         & \parlam\ 256k & 13.37 & 50.70 & 0.82 & 0.79 & 10.26 & 46.86 & 0.80 & 0.77 \\
         & \parlam\ 32k & 12.68 & 49.77 & 0.81 & 0.78 & 10.50 & 46.72 & 0.79 & 0.76 \\
        \cdashlinelr{1-10}
        pt-fr & \textsc{NLLB 3.3B}  & 40.85 & 64.94 & 0.87 & 0.86 & 29.39 & 57.41 & 0.84 & 0.85 \\
         & \textsc{TowerBase 7B} & 36.52 & 62.44 & 0.85 & 0.85 & 29.39 & 57.41 & 0.84 & 0.85 \\
         \cdashlinelr{2-10}
         & \parlam\ 128k & 33.25 & 59.78 & 0.83 & 0.83 & 23.91 & 52.93 & 0.80 & 0.81 \\
         & \parlam\ 256k & 33.80 & 59.69 & 0.83 & 0.82 & 24.72 & 53.34 & 0.81 & 0.81 \\
         & \parlam\ 32k & 32.60 & 58.97 & 0.82 & 0.82 & 24.11 & 52.80 & 0.80 & 0.80 \\
        \cdashlinelr{1-10}
        pt-gl & \textsc{NLLB 3.3B}  & 31.12 & 57.92 & 0.88 & 0.83 & 32.55 & 59.00 & 0.87 & 0.82 \\
        \cdashlinelr{2-10}
         & \parlam\ 128k & 28.83 & 56.91 & 0.87 & 0.82 & 28.27 & 56.48 & 0.85 & 0.81 \\
         & \parlam\ 256k & 28.58 & 56.52 & 0.87 & 0.82 & 28.54 & 56.57 & 0.85 & 0.81 \\
         & \parlam\ 32k & 28.64 & 56.61 & 0.87 & 0.82 & 28.01 & 56.32 & 0.85 & 0.81 \\
        \cdashlinelr{1-10}
        pt-it & \textsc{NLLB 3.3B}  & 26.42 & 55.44 & 0.88 & 0.85 & 31.19 & 59.11 & 0.87 & 0.85 \\
         & \textsc{TowerBase 7B} & 22.31 & 52.69 & 0.85 & 0.85 & 31.19 & 59.11 & 0.87 & 0.85 \\
         \cdashlinelr{2-10}
         & \parlam\ 128k & 24.06 & 53.75 & 0.86 & 0.84 & 26.97 & 56.30 & 0.85 & 0.83 \\
         & \parlam\ 256k & 24.24 & 53.75 & 0.86 & 0.84 & 27.46 & 56.52 & 0.85 & 0.83 \\
         & \parlam\ 32k & 23.67 & 53.46 & 0.85 & 0.83 & 27.60 & 56.49 & 0.85 & 0.83 \\

        \bottomrule
    \end{tabular}
\end{table*}

\end{document}